\documentclass[11pt, a4paper]{lumia}

\usepackage[sort&compress]{natbib}
\usepackage{longtable}  
\usepackage{array}      
\usepackage{xltabular}
\usepackage{arydshln}
\usepackage[commandnameprefix=always]{changes}
\usepackage{booktabs}    
\setheadertext{title}
\usepackage[most,skins,theorems]{tcolorbox} 

\newtcolorbox{promptbox}[2][]{
  colback=gray!5,       
  colframe=black!70,    
  title=\textbf{#2},     
  fonttitle=\bfseries\small,
  fontupper=\footnotesize\ttfamily, 
  left=3pt, right=3pt, top=3pt, bottom=3pt, 
  boxrule=0.8pt,
  sharp corners,         
  #1
}

\renewcommand{\today}{2026-05}

\usepackage{xspace}

\theoremstyle{plain}

\newtheorem*{proposition*}{Proposition}

\theoremstyle{definition}

\theoremstyle{definition}

\def\eqref#1{equation~\ref{#1}}

\usepackage[utf8]{inputenc}     
\usepackage[T1]{fontenc}        
\usepackage{microtype}          

\usepackage{xcolor}             
\usepackage[dvipsnames]{xcolor} 
\usepackage{graphicx}           
\usepackage{tikz}               
\usepackage[edges]{forest}      

\usepackage{hyperref}           
\usepackage{url}                
\usepackage{xurl}               

\usepackage{tabularx}           
\usepackage{array}              
\usepackage{booktabs}           
\usepackage{longtable}          
\usepackage{multirow}           
\usepackage{makecell}           
\usepackage[table]{xcolor}      
\usepackage{colortbl}           
\usepackage{ragged2e}           

\usepackage{amsmath}            
\usepackage{mathtools}          
\usepackage{amsfonts}           
\usepackage{amssymb}            
\usepackage{amsthm}             
\usepackage{nicefrac}           

\usepackage{algorithm}          
\usepackage{algorithmicx}       
\usepackage{algpseudocode}      
\usepackage{listings}           

\usepackage{subcaption}         
\usepackage{caption}            
\usepackage{wrapfig}            
\usepackage[export]{adjustbox}  

\usepackage{soul}               
\usepackage{changes}            
\usepackage{xspace}             
\usepackage[normalem]{ulem}     
\usepackage{CJKutf8}            

\usepackage{enumitem}           
\usepackage{pifont}             

\usepackage[most,skins,theorems]{tcolorbox} 
\usepackage[tikz]{bclogo}       
\usepackage[framemethod=tikz]{mdframed} 

\usepackage{lipsum}             
\usepackage{tocloft}            
\usepackage{fontawesome5}       
\usepackage{afterpage}          
\usepackage{bbding}             
\usepackage{epigraph}           
\usepackage{minitoc}            
\usepackage{multicol}           
\usepackage{textgreek}          

\newcolumntype{P}[1]{>{\RaggedRight\arraybackslash}p{#1}}

\definecolor{darkblue}{rgb}{0, 0, 0.5}
\definecolor{uclablue}{RGB}{39, 116, 174}
\definecolor{bigaired}{RGB}{156, 0, 0}
\definecolor{myblue}{HTML}{598BE7}
\definecolor{mildblue}{RGB}{31,119,180}
\definecolor{sectionblue}{RGB}{70, 130, 180}
\definecolor{methodblue}{RGB}{0, 150, 136}
\definecolor{bgblue}{RGB}{245,243,253}
\definecolor{ttblue}{RGB}{91,194,224}
\definecolor{mygreen}{rgb}{0.64, 0.56, 0.88}
\definecolor{myyellow}{rgb}{0.68, 0.6, 0.1}
\definecolor{fancygreen}{rgb}{0.33, 0.68, 0.20}
\definecolor{salmon}{rgb}{0.94, 0.52, 0.49}
\definecolor{tablegreen}{rgb}{0.82, 0.94, 0.75}
\definecolor{tableblue}{rgb}{0.81, 0.90, 0.94}
\definecolor{tablered}{rgb}{0.97, 0.85, 0.85}
\definecolor{tableorange}{rgb}{0.96, 0.85, 0.81}
\definecolor{myorange}{rgb}{1.0, 0.49, 0.0}
\definecolor{tlgreen}{rgb}{0.33, 0.68, 0.20}
\definecolor{darkgreen}{RGB}{0,100,0}
\definecolor{darkred}{RGB}{200, 0, 0}
\definecolor{lightblue}{RGB}{220,235,250}
\definecolor{customyellow}{HTML}{FFFACD}
\definecolor{refinegreen}{RGB}{0, 128, 75}
\definecolor{scoregreen}{RGB}{34, 139, 34}
\definecolor{hidden-blue}{RGB}{194,232,247}
\definecolor{hidden-black}{RGB}{20,68,106}
\definecolor{yes}{HTML}{C6EFCE}
\definecolor{no}{HTML}{FFC7CE}
\definecolor{partial}{HTML}{FFEB9C}
\definecolor{external}{HTML}{D9E1F2}
\definecolor{hdr}{HTML}{F2F2F2}
\definecolor{GRPOrow}{gray}{0.96}
\definecolor{FlowRLrow}{RGB}{225,236,255}
\definecolor{FlowBlue}{RGB}{80,120,210}
\definecolor{GRPOGray}{gray}{0.35}

\hypersetup{
    colorlinks=true, 
    citecolor=uclablue, 
    linkcolor=bigaired,
    urlcolor=darkblue
}

\setlist[itemize]{leftmargin=20pt, noitemsep, topsep=0pt}

\NewDocumentCommand{\kaiyan}{mO{}}{\textcolor{purple}{\textsuperscript{\textit{kaiyan}}\textsf{\textbf{\small[#1]}}}}
\NewDocumentCommand{\yuxin}{mO{}}{\textcolor{cyan}{\textsuperscript{\textit{yuxin}}\textsf{\textbf{\small[#1]}}}}
\NewDocumentCommand{\bx}{mO{}}{\textcolor{green}{\textsuperscript{\textit{bx}}\textsf{\textbf{\small[#1]}}}}
\NewDocumentCommand{\at}{mO{}}{\textcolor{red}{\textsuperscript{\textit{AT}}\textsf{\textbf{\small[#1]}}}}
\NewDocumentCommand{\re}{mO{}}{\textcolor{blue}{\textsuperscript{\textit{RE}}\textsf{\textbf{\small[#1]}}}}
\NewDocumentCommand{\ybsun}{mO{}}{\textcolor{magenta}{\textsuperscript{\textit{youbang}}\textsf{\textbf{\small[#1]}}}}
\NewDocumentCommand{\runze}{mO{}}{\textcolor{orange}{\textsuperscript{\textit{runze}}\textsf{\textbf{\small[#1]}}}}
\NewDocumentCommand{\add}{mO{}}{\textcolor{darkgreen}{\textsuperscript{\textit{Maybe Consider Discuss}}\textsf{\textbf{[#1]}}}}

\newcommand{\cmark}{\textcolor{darkgreen}{\boldmath$\checkmark$}}
\newcommand{\xmark}{\textcolor{darkred}{\boldmath$\times$}}

\newenvironment{itemize*}%
 {\leftmargini=10pt\begin{itemize}%
  \setlength{\itemsep}{0pt}%
  \setlength{\parskip}{0pt}%
  }%
 {\end{itemize}}

\newenvironment{enumerate*}%
 {\begin{enumerate}%
  \setlength{\itemsep}{0pt}%
  \setlength{\parskip}{0pt}}%
 {\end{enumerate}}

\newcommand{\cellstatus}[1]{%
  \begingroup
  \StrTrim{#1}[\statusval]%
  \IfStrEq{\statusval}{Yes}{\cellcolor{yes}\cmark}{}%
  \IfStrEq{\statusval}{No}{\cellcolor{no}\xmark}{}%
  \IfBeginWith{\statusval}{Yes (}{\cellcolor{yes}\cmark~\textit{\statusval\unskip}}{}%
  \IfStrEq{\statusval}{Partial}{\cellcolor{partial}\textbf{Partial}}{}%
  \IfStrEq{\statusval}{External}{\cellcolor{external}\textbf{External}}{}%
  \endgroup
}

\newtcolorbox{myboxi}[1][]{
  breakable,
  title=#1,
  colback=red!5,
  colbacktitle=red!5,
  coltitle=black,
  fonttitle=\bfseries,
  bottomrule=0pt,
  toprule=0pt,
  leftrule=2pt,
  rightrule=2pt,
  titlerule=0pt,
  arc=0pt,
  outer arc=0pt,
  colframe=red,
}

\newtcolorbox{myboxnote}[1][]{
  breakable,
  title=#1,
  colback=orange!0,
  colbacktitle=orange!0,
  coltitle=black,
  fonttitle=\bfseries,
  bottomrule=0pt,
  toprule=0pt,
  leftrule=2pt,
  rightrule=2pt,
  titlerule=0pt,
  arc=0pt,
  outer arc=0pt,
  colframe=orange,
}

\newtcolorbox{myboxii}[1][]{
  breakable,
  freelance,
  title=#1,
  colback=white,
  colbacktitle=white,
  coltitle=black,
  fonttitle=\bfseries,
  bottomrule=0pt,
  boxrule=0pt,
  colframe=white,
  overlay unbroken and first={
  \draw[red!75!black,line width=3pt]
    ([xshift=5pt]frame.north west) -- 
    (frame.north west) -- 
    (frame.south west);
  \draw[red!75!black,line width=3pt]
    ([xshift=-5pt]frame.north east) -- 
    (frame.north east) -- 
    (frame.south east);
  },
  overlay unbroken app={
  \draw[red!75!black,line width=3pt,line cap=rect]
    (frame.south west) -- 
    ([xshift=5pt]frame.south west);
  \draw[red!75!black,line width=3pt,line cap=rect]
    (frame.south east) -- 
    ([xshift=-5pt]frame.south east);
  },
  overlay middle and last={
  \draw[red!75!black,line width=3pt]
    (frame.north west) -- 
    (frame.south west);
  \draw[red!75!black,line width=3pt]
    (frame.north east) -- 
    (frame.south east);
  },
  overlay last app={
  \draw[red!75!black,line width=3pt,line cap=rect]
    (frame.south west) --
    ([xshift=5pt]frame.south west);
  \draw[red!75!black,line width=3pt,line cap=rect]
    (frame.south east) --
    ([xshift=-5pt]frame.south east);
  },
}

\tcbset{
  takeawaysbox/.style={
    title=Takeaways,
    colback=lightblue!80,
    colframe=black,
    fonttitle=\bfseries\small,
    coltitle=white,
    colbacktitle=black,
    enhanced,
    attach boxed title to top left={xshift=2.5mm,yshift=-2.5mm},
    boxed title style={rounded corners, size=small, colframe=black, colback=black},
    width=\linewidth,
    arc=3.5mm
  }
}

\mdfdefinestyle{mystyle}{%
  rightline=true,
  innerleftmargin=10,
  innerrightmargin=10,
  outerlinewidth=3pt,
  topline=false,
  rightline=true,
  bottomline=false,
  skipabove=\topsep,
  skipbelow=\topsep
}

\tikzset{%
    every node/.style={font=\tiny},
    parent/.style =          {align=center,text width=2cm,rounded corners=3pt, line width=0.3mm, fill=gray!10,draw=gray!80},
    child/.style =           {align=center,text width=2.0cm,rounded corners=3pt, fill=blue!10,draw=blue!80,line width=0.3mm},
    grandchild/.style =      {align=center,text width=2cm,rounded corners=3pt},
    greatgrandchild/.style = {align=center,text width=1.5cm,rounded corners=3pt},
    greatgrandchild2/.style = {align=center,text width=1.5cm,rounded corners=3pt},    
    referenceblock/.style =  {align=center,text width=1.5cm,rounded corners=2pt},
    pretrain/.style =           {align=center,text width=2.0cm,rounded corners=3pt, fill=blue!10,draw=blue!80,line width=0.3mm},   
    pretrain_work/.style =           {align=center, text width=8.5cm,rounded corners=3pt, fill=blue!10,draw=blue!0,line width=0.3mm},  
    template/.style =           {align=center,text width=2.0cm,rounded corners=3pt, fill=red!10,draw=red!80,line width=0.3mm},   
    template_work/.style =           {align=center,text width=8.5cm,rounded corners=3pt, fill=red!10,draw=red!0,line width=0.3mm},    
    answer/.style =           {align=center,text width=2.0cm,rounded corners=3pt, fill= cyan!10,draw= cyan!80,line width=0.3mm},   
    answer_work/.style =           {align=center,text width=8.5cm,rounded corners=3pt, fill= cyan!10,draw= cyan!0,line width=0.3mm},      
    multiple/.style =           {align=center,text width=2.0cm,rounded corners=3pt, fill= orange!10,draw= orange!80,line width=0.3mm},   
    multiple_work/.style =           {align=center,text width=8.5cm,rounded corners=3pt, fill= orange!10,draw= orange!0,line width=0.3mm},        
    tuning/.style =           {align=center,text width=2.0cm,rounded corners=3pt, fill= magenta!10,draw= magenta!80,line width=0.3mm},   
    tuning_work/.style =           {align=center,text width=8.5cm,rounded corners=3pt, fill= magenta!10,draw= magenta!0,line width=0.3mm},          
}

\newcommand{\lstbg}[3][0pt]{{\fboxsep#1\colorbox{#2}{\strut #3}}}

\lstdefinelanguage{diff}{
  basicstyle=\ttfamily\small,
  morecomment=[f][\lstbg{red!20}]-,
  morecomment=[f][\lstbg{green!20}]+,
}

\lstdefinelanguage{diffpython}{
  language=diff,
  morekeywords={def, if, else, for, while, return, import, from, as, class, with, try, except, finally, raise, lambda, and, or, not, in, is, None, True, False},
  morecomment=[l]{\#},
  morestring=[b]",
  morestring=[b]',
}

\title{InteractWeb-Bench: Can Multimodal Agent Escape Blind Execution in Interactive Website Generation?}

\author{
    \scriptsize
    Qiyao Wang$^{1,2,*}$\quad 
    Haoran Hu$^{3,*}$\quad 
    Longze Chen$^{1,2}$\quad 
    Hongbo Wang$^{3}$\quad 
    Hamid Alinejad-Rokny$^{4}$\quad
    Yuan Lin$^{3\dagger}$\quad
    Min Yang$^{1,5\dagger}$ \\
    $^1$~\raisebox{-0.5ex}{\includegraphics[height=2.9ex]{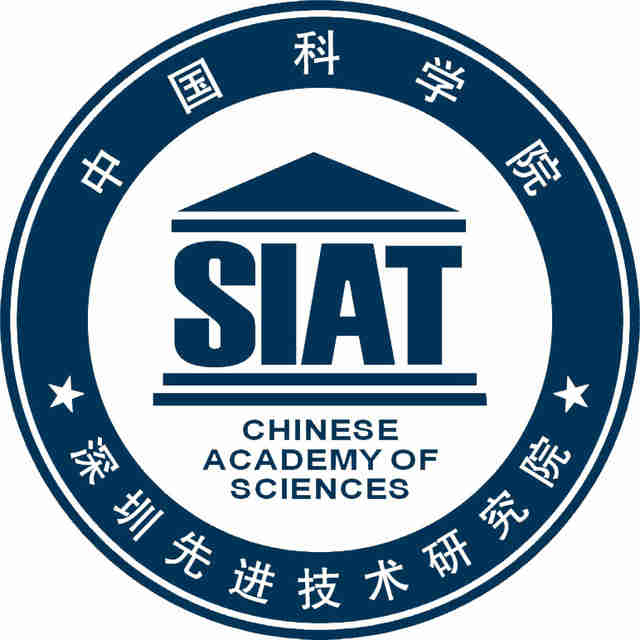}}~Shenzhen Institute of Advanced Technology, Chinese Academy of Sciences \quad
    $^2$~\raisebox{-0.5ex}{\includegraphics[height=2.9ex]{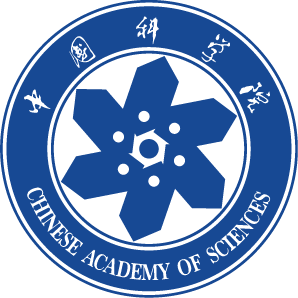}}~University of Chinese Academy of Sciences\\
    $^3$~\raisebox{-0.5ex}{\includegraphics[height=2.9ex]{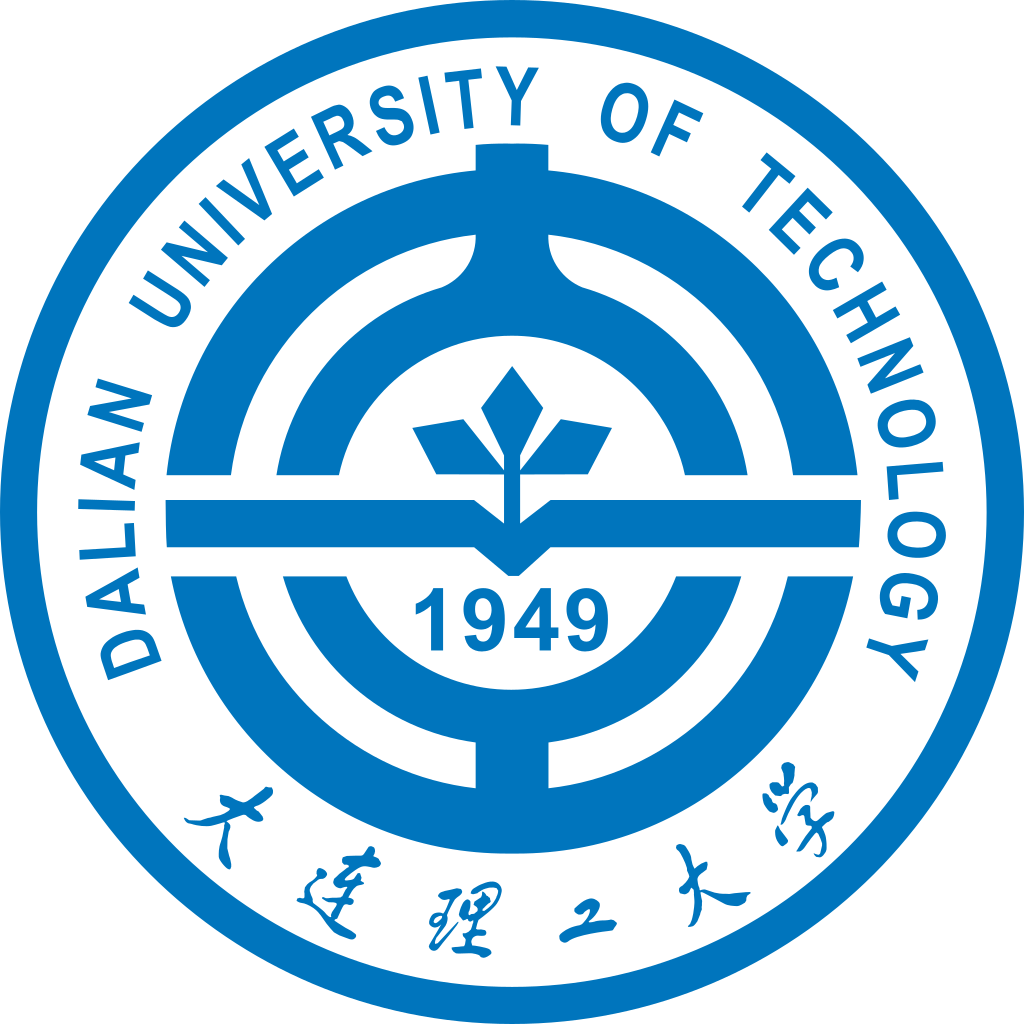}}~Dalian University of Technology\quad
    $^4$~\raisebox{-0.5ex}{\includegraphics[height=2.9ex]{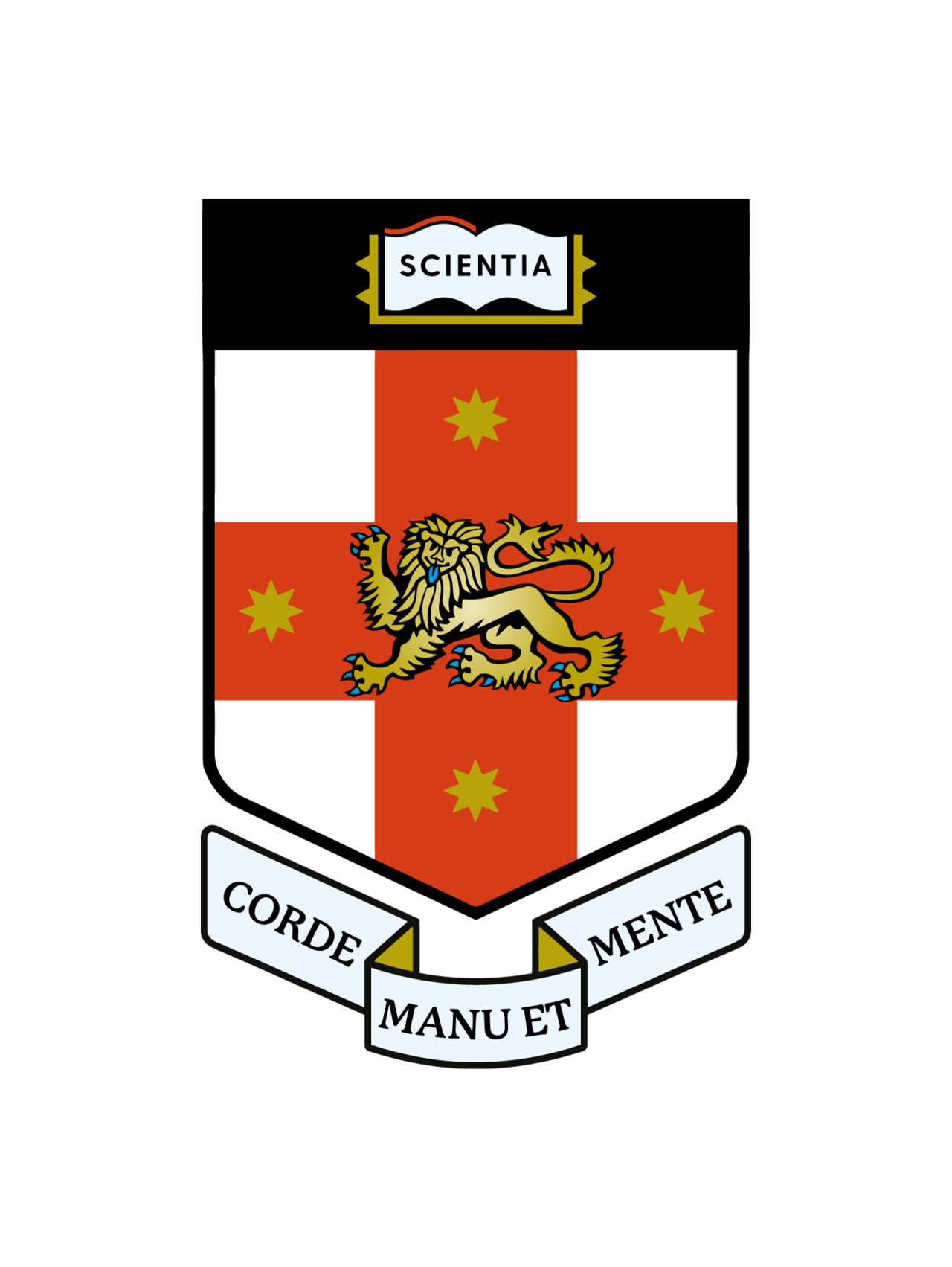}}~UNSW Sydney\quad
    $^5$~\raisebox{-0.5ex}{\includegraphics[height=2.9ex]{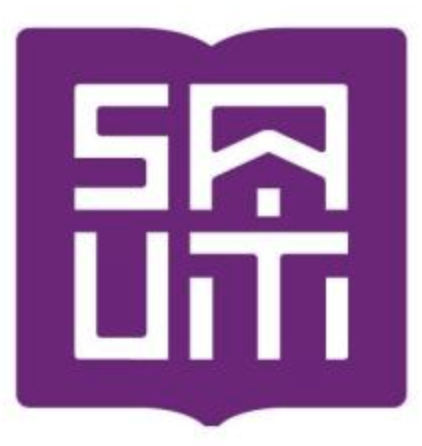}}~Shenzhen University of Advanced Technology \\
    \faEnvelope[regular]~\texttt{wangqiyao25@mails.ucas.ac.cn}  \quad
    \faEnvelope[regular]~\texttt{zhlin@dlut.edu.cn}  \quad
    \faEnvelope[regular]~\texttt{min.yang@siat.ac.cn}  \\
    \faHome~\href{https://interactweb-bench.wangqiyao.me/}{Website} \quad \faGithub~\href{https://github.com/AIforIP/InteractWeb-Bench}{InteractWeb-Bench} \quad
    $^*$ Equal Contribution. \quad
    $^\dagger$ Corresponding Authors. 
}

\begin{abstract}
With the advancement of multimodal large language models~(MLLMs) and coding agents, the website development has shifted from manual programming to agent-based project-level code synthesis. 
Existing benchmarks rely on idealized assumptions, especially for well-structured, information-rich inputs and static execution settings. 
In contrast, real-world development is constrained by a critical bottleneck: the semantic misalignment between ambiguous, low-quality instructions from non-expert users and model understanding, which results in a failure mode that we term \textbf{\textit{blind execution}}.
To address this gap, we introduce \textbf{InteractWeb-Bench}, the first multimodal interactive benchmark for website generation under non-expert low-code user conditions. 
InteractWeb-Bench introduces four types of user agents and persona-driven instruction perturbations to systematically simulate diverse user behaviors, including ambiguity, redundancy, and contradiction, grounded in requirement engineering defect taxonomies.
We develop an interactive execution environment for agents, featuring a unified action space comprising \textit{Clarify}, \textit{Implement}, \textit{Verify}, and \textit{Submit}, enabling iterative intent refinement, code synthesis, and visual feedback–based validation.
Extensive experiments and analysis reveal that frontier MLLM-based agents remain trapped in blind execution, exposing limitations in intent recognition and adaptive interaction.
\end{abstract}

\begin{document}
\maketitle
\section{Introduction}

With the advancement of multimodal large language models (MLLMs) and coding agents~\citep{Claude3S, singh2025openaigpt5card}, the paradigm of website development is evolving from traditional \textit{manual code authoring} towards \textit{natural language-guided code synthesis}~\citep{lu2025webgenbenchevaluatingllmsgenerating}. 
This substantially reduces the barriers to professional software development, allowing non-expert users to directly participate through natural language instructions~\citep{hou2024large}.
In an ideal scenario, the agent should accurately parse user requirements and end-to-end perform code synthesis to generate webpages. 
However, the inherent ambiguity and low-code nature of instructions from non-expert users often lead to a significant semantic gap between their true intent and model's understanding~\citep{zamfirescu2023why} in practice. 
This constitutes one of the primary bottlenecks for website generation in moving toward general real-world deployment~\citep{jimenez2024swebench}.

Existing website generation benchmarks are confined to an idealized evaluation setting~\citep{lu2025webgenbenchevaluatingllmsgenerating}.
In these environments, the instructions provided are often standardized, with clear structures, rigorous logic, and comprehensive frontend details.
However, real-world interactive scenarios are far from perfect. 
Constrained by a lack of domain knowledge, especially in programming, the initial requirements provided by non-expert users typically exhibit highly ambiguous and fragmented characteristics.
Furthermore, diverse user groups exhibit drastically different expression styles, such as minimalist abstract descriptions, divergent expressions containing substantial redundant information, or even conflicting requirements that violate technical common sense~\citep{luo2025clarifymtbenchbenchmarkingimprovingmultiturn}.

Faced with such low-quality, high-variance instructions, existing agent commonly fall into the trap of \textit{\textbf{Blind Execution}}. 
It passively adapts flawed queries, lacks the ability to effectively infer the user’s true intent, and cannot proactively request clarification~\citep{wu2025codelanguagemodelslearn} or employ dynamic reasoning-and-acting strategies ~\citep{yao2023react} to bridge missing information gaps.
These models typically default to bypassing necessary intent verification, directly synthesizing code from incomplete or conflicting instructions. 
This unidirectional execution strategy directly leads to elevated task failure rates, UI rendering anomalies, and high-frequency functional hallucinations~\citep{ji2023survey}.

To escape the trap of {{blind execution}}, an agent must possess robust capabilities in \textit{dynamic intent recognition}. 
While recent studies emphasize proactive conversational assistance~\citep{lu2024proactiveagentshiftingllm} and intent clarification in language and code models~\citep{luo2025clarifymtbenchbenchmarkingimprovingmultiturn, wu2025codelanguagemodelslearn}, it is fundamentally insufficient to confine interactions to text-centric dialogues in the highly interdependent and visually intensive domain of frontend engineering.
The non-expert user requirements are rarely articulated perfectly upfront, agents must make interleaved decisions, dynamically switching between clarifying ambiguous user intent and verifying visual webpage outputs in an executable environment.

\begin{figure}[!t]
    \includegraphics[width=\linewidth]{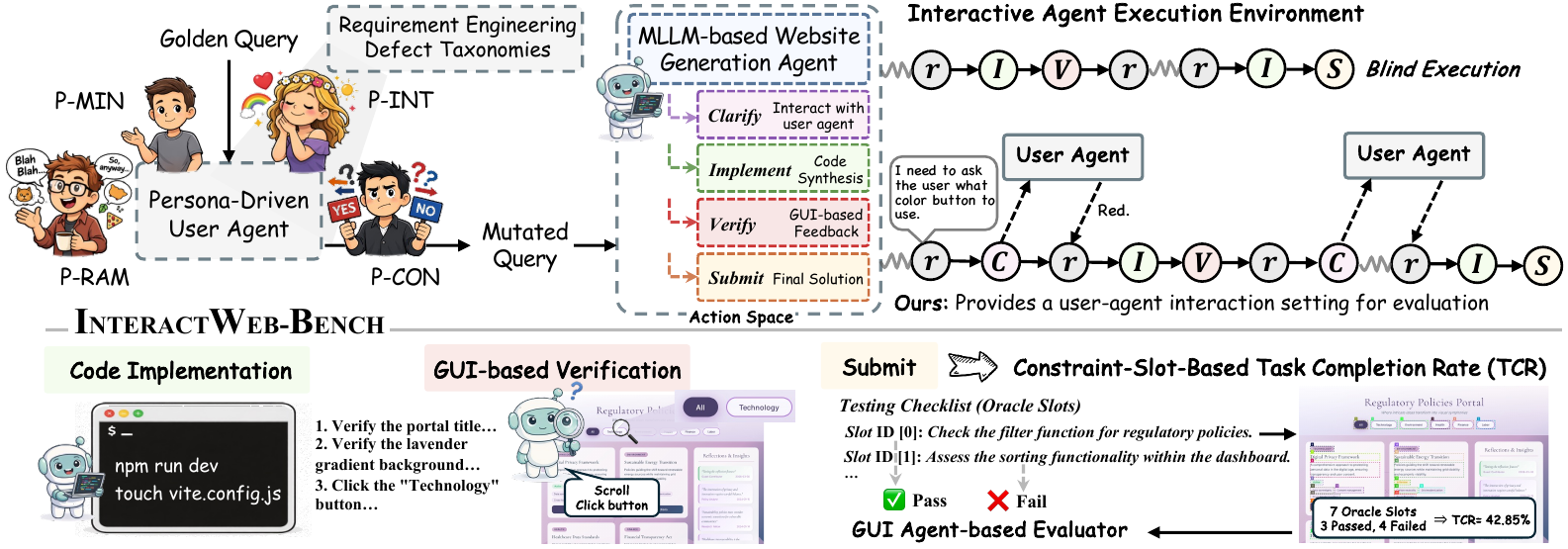} 
    \caption{The overall architecture of InteractWeb-Bench. Where the top-right circles denote actions in an illustrative trajectory: \textit{r} (reasoning), \textit{C} (Clarify), \textit{I} (Implement), \textit{V} (Verify), and \textit{S} (Submit).}
    \label{fig:framework_overview}
\end{figure}

To address these limitations, we introduce \textbf{InteractWeb-Bench}, the first multimodal interactive benchmark for non-expert user low-code website generation, as illustrated in Figure~\ref{fig:framework_overview}. 
It is designed to evaluate whether agents can move beyond blind execution by dynamically recognizing user intent and interacting with users. 
To authentically simulate the complexity of user instructions, we construct four typical user personas (i.e., P-MIN, P-RAM, P-INT, and P-CON), grounded in standard Requirement Engineering defect taxonomies~\citep{van2009requirements,10.1145/3715109}, to inject \textit{intent traps} serving as stress tests. 
Specifically, to enable agents to navigate these highly uncertain inputs, we designs a unified multi-path action space comprising natural language \textit{clarification (Clarify)}, \textit{code synthesis (Implement)}, \textit{visual verification (Verify)}, and \textit{task submission (Submit)}. 
Within this environment, the agent dynamically selects any of these actions based on the real-time state. 
It freely toggles between eliciting requirements, writing code, and visually inspecting the rendered UI. Finally, it delivers the completed website to our designed constraint--slot-based evaluating procedure.
We envision our InteractWeb-Bench could catalyze the evolution of MLLM-based website generation agents from passive instruction followers to intent-aligned collaborators capable of escaping blind execution. \textbf{Our main contributions are as follows:}

\begin{itemize}
    \item We focus on the critical bottleneck of \textit{Blind Execution} in website generation agents. To address the limitations of idealized static benchmarks, we incorporate the \textit{user agents} and \textit{persona-driven instruction mutation} to better capture the ambiguity and heterogeneity of user intents.    
    \item To the best of our knowledge, our \textbf{InteractWeb-Bench} is the first benchmark to evaluate MLLM-based agents' ability to interact with users and satisfy their needs in website generation, by endowing the models with a unified multi-path action space (\textit{Clarify}, \textit{Implement}, \textit{Verify}, \textit{Submit}).
   
   \item We conduct extensive experiments on various frontier MLLM-based agents and provide in-depth analyses of their behavioral patterns, especially for user–agent interaction, intent recognition, and inherent task-solving capabilities. 
    Our findings reveal key factors for improving website generation agents and better aligning their behavior with the needs of non-expert users.
\end{itemize}

\section{Related Work}

\paragraph{Website Generation Benchmarks.}
The application of MLLMs and coding agents in software engineering has advanced automated code generation~\citep{chen2021evaluatinglargelanguagemodels,austin2021programsynthesislargelanguage}, with increasing research focusing on website generation.
Early work formulate this task as a static vision-to-code translation problem. 
For instance, Design2Code~\citep{si-etal-2025-design2code} and Web2Code~\citep{NEURIPS2024_cb66be28} assess the ability to convert high-fidelity webpage screenshots or design specifications into static HTML/CSS implementations.
WebGen-Bench~\citep{lu2025webgenbenchevaluatingllmsgenerating} advanced the paradigm by evaluating the generation of multi-page web applications from scratch based on comprehensive textual instructions. 
However, existing benchmarks rely on highly standardized and comprehensive instructions, assuming the initial query fully captures user preferences and failing to evaluate models under the high-variance, ambiguous, and contradictory inputs typical of real-world scenarios.
In contrast, our InteractWeb-Bench explicitly targets this gap by introducing persona-driven instruction variance to test models under realistic, noisy conditions.

\paragraph{Intent Clarification and User Interaction.}
Recent work actively explores proactive intent clarification in response to ambiguous instructions. 
ClarifyCoder~\citep{wu2025codelanguagemodelslearn} enables models to request missing specifications via instruction-tuning.
HumanEvalComm~\citep{10.1145/3715109} and ClarifyMT-Bench~\citep{luo2025clarifymtbenchbenchmarkingimprovingmultiturn} introduce flawed requirements and noisy user personas to evaluate robustness under incomplete, ambiguous or noisy instructions.
In website generation, Persona2Web~\citep{kim2026persona2webbenchmarkingpersonalizedweb} have explored ambiguous queries, but they restrict agents to passively inferring intents from browsing histories rather than actively communicating. 
While these work realize the importance of requirement elicitation and ambiguity resolution, they remain fundamentally confined to either pure textual content or passive inference. 
In the highly visual field of website frontend engineering, agents must leverage iterative visual rendering feedback to enhance their understanding.
Our InteractWeb-Bench bridges this gap by integrating language requirement elicitation with visual verification.

\paragraph{Interactive Agents and Execution Environments.}
The evaluation paradigm of LLMs has shifted from static code generation to interactive agents operating within environments~\citep{yao2023react}. 
General-purpose code agent frameworks, such as SWE-agent~\citep{NEURIPS2024_5a7c9475} and OpenHands~\citep{wang2025openhandsopenplatformai}, interact with bash terminals and code editors to obtain execution feedback (e.g., compiler errors, test outputs). 
Correspondingly, benchmarks like SWE-bench~\citep{jimenez2024swebench} and InterCode~\citep{yang2023intercode} evaluate agents based on their ability to use this feedback over multiple turns to resolve real-world GitHub issues. 
However, these environments focus primarily on passing objective, predefined functional tests.
They largely ignore the human-AI collaborative process, failing to assess whether an agent can proactively elicit and satisfy unstated user needs. 
Because non-expert users frequently provide flawed or incomplete instructions, an agent must move beyond mere code execution and escape the trap of \textit{blind execution}.
This motivates our InteractWeb-Bench, which immerses the agent in a comprehensive multi-path action space (\textit{Clarify, Implement, Verify, Submit}), enabling it to autonomously toggle between textual requirement clarification, code synthesis, and GUI-based visual debugging to fulfill the user's true intent.

\section{InteractWeb-Bench}

To evaluate whether MLLM-based agents can escape the \textit{blind execution} and navigate the ambiguity of real-world user requirements in website generation, we propose \textbf{InteractWeb-Bench}, an \underline{Interact}ive \underline{Web}site generation \underline{Bench}mark.
As illustrated in Figure~\ref{fig:framework_overview} and Algorithm~\ref{algorithm1}, it incorporates user simulation modules to evaluate agents in natural language-driven website generation tasks.
Specifically, our framework consists of two core components: \textbf{(1) Persona-Driven User Agent Module}, comprising four carefully designed user agents that simulate diverse human inputs and introduce targeted intent traps (See Section~\ref{sec:user}); 
and \textbf{(2) Interactive Execution Environment}, featuring a unified action space that empowers the agent to autonomously toggle among language clarification (\textit{Clarify}), code synthesis (\textit{Implement}), GUI-based visual verification (\textit{Verify}), and task submission (\textit{Submit}) (See Section~\ref{sec:env}). 
Unlike prior benchmarks, InteractWeb-Bench is designed to evaluate agents' capability in website generation, focusing on requirements elicitation, cross-modal comprehension, code synthesis, and iterative refinement.

\subsection{Persona-Driven User Agent Module}
\label{sec:user}

\begin{wrapfigure}{r}{0.451\textwidth}
\vspace{-18pt}
\refstepcounter{algorithm}
\label{algorithm1}
\begin{tcolorbox}[
    title=Algorithm 1: InteractWeb-Bench Evaluation,
    colback=white,
    colframe=black,
    boxrule=0.5pt,
    arc=2pt,
    left=0pt, right=0pt, top=0pt, bottom=0pt,
    fonttitle=\bfseries\footnotesize,
]
\footnotesize
\begin{algorithmic}[1]
\State \textbf{// Persona-driven Initialization}
\State Instruction $\mathcal{I} \gets O_p(\mathcal{I}^*)$;
\State Initialize user agent $U_p$ and execution agent $\pi_{\mathcal{M}}$;
\State Initialize environment state $s_0$, Timestep $t \gets 0$;
\State \textbf{// Interactive Agent Execution}
\While{not terminated}
    \State $a_t \sim \pi_{\mathcal{M}}(s_t)$.
    \State $a_t \in \{{\small\textit{clarify, implement, verify, submit}}\}$.
    \State Execute $a_t$ via (one-choice):
    \State \hspace{1em} Clarify $\rightarrow$ Interact with $U_p$;
    \State \hspace{1em} Implement $\rightarrow$ Generate code;
    \State \hspace{1em} Verify $\rightarrow$ GUI inspection;
    \State \hspace{1em} Submit $\rightarrow$ Terminate proactively.
\EndWhile
\State \textbf{// Final Evaluation}
\State Compute final task completion rate score.
\end{algorithmic}
\end{tcolorbox}
\vspace{-14pt}
\end{wrapfigure}
\paragraph{Persona-Driven Instruction Mutation.}
In practical software engineering, initial user requirements are frequently underspecified or articulated with high variability.
To systematically simulate these real-world communication barriers, our persona design is theoretically grounded in standard Requirement Engineering (RE) defect taxonomies~\citep{van2009requirements,10.1145/3715109} (e.g., incompleteness, ambiguity, and inconsistency) and Grice's Maxims of Conversation~\citep{grice1975logic}.
We formalize these theoretical requirement flaws into four distinct user personas. 
Let $\mathcal{I}^*$ denote the original, unambiguous golden instruction.
By defining these four personas, we apply a persona-specific metamorphic operator $O_{\textit{\text{persona}}}$ to generate the perturbed instruction $\mathcal{I}_{\textit{\text{mutated}}} = O_{\textit{\text{persona}}}(\mathcal{I}^*)$~\citep{chen2020metamorphictestingnewapproach}, emulating specific types of human users:
\begin{itemize}
    \item \textbf{Minimalist User Agent (P-MIN):} Represents \textit{Requirement Incompleteness} (violating the Maxim of Quantity), which simulates a user who provides extremely sparse details.
    Specifically, we apply an abstraction operator to remove secondary constraints (e.g., colors or layout-related terminology), retaining only the core verb–object functional request.
    This rigorously tests the agent's ability to proactively elicit missing requirements and apply sensible design defaults.
    
    \item \textbf{Rambling User Agent (P-RAM):} Represents \textit{Low Signal-to-Noise Ratio} (violating the Maxim of Relation), which simulates a highly talkative and unfocused user. Specifically, we utilize a noise-injection operator to bury core functional requirements within substantial non-functional conversational digressions and irrelevant background context, challenging the agent's capacity for noise filtering, robust information extraction, and instruction tracking amidst distracting inputs.
    
    \item \textbf{Intuitive User Agent (P-INT):} Represents \textit{Requirement Ambiguity and Informality} (violating the Maxim of Manner), which represents a non-technical user highly focused on sensory experiences and aesthetics. Specifically, we employ a cross-modal semantic shift to translate precise frontend engineering terminology into abstract sensory metaphors and emotional modifiers, testing the agent's visual and textual alignment capabilities.
    
    \item \textbf{Conflicting User Agent (P-CON):} Represents 
    \textit{Requirement Contradiction} (violating the Maxim of Quality), which simulates a confused or demanding user who provides mutually exclusive constraints. 
    Specifically, we introduce logical inconsistencies and frontend paradoxes into the interdependent requirements, evaluating whether the agent can accurately identify these invalid propositions and proactively question the user to resolve the conflict back to the original truth.
\end{itemize}

\paragraph{User Agent Response Design.}
The second stage of user agent engagement involves dynamic, multi-turn communication with execution agent. 
The execution agent must proactively elicit missing information and resolve conflicts under ambiguous or incomplete initial instructions, while avoiding shortcut strategies that directly solicit complete solutions from the user.
We employ a two-stage {{retrieval and containment}} pipeline for user agent to prevent the simulated user from inadvertently revealing the entire unperturbed golden instruction upon a single query. 

First, the user agent analyzes the execution agent's clarification request and retrieves the internal ground-truth specification.
By extracting only the explicitly requested details and withholding all unasked requirements, the system effectively prevents information leakage. 
Second, through {persona-conditioned response generation}, the isolated factual snippet is processed via a persona-specific prompt filter, seamlessly weaving the oracle data into the assigned user identity's conversational style. 
For instance, if asked about missing background color, the P-RAM user agent will embed the exact true requirements within substantial irrelevant conversational noise.
Additionally, if the execution agent correctly identifies a technical paradox, the P-CON user agent is instructed to briefly defend its original contradictory stance before ultimately yielding the accurate baseline truth.

\subsection{Interactive Agent Framework and Execution Environment}
\label{sec:env}
We instantiate the evaluated MLLM as an autonomous website generation agent within an interactive execution environment. 
Through this setup, the agent iteratively engages with user instructions. 
It structures the agent's behavior through a unified action space and a multi-modal feedback loop, testing the holistic capabilities of requirement elicitation, code implementation, and visual comprehension.

\paragraph{Action Space ($\mathcal{A}$) and Autonomous Trajectory Dynamics.} 
Unlike fixed workflows, the agent operates in our environment with a fully autonomous, non-linear manner. 
At any step $t$, based on its current observation and internal reasoning, the agent dynamically selects an action from a predefined discrete action space $\mathcal{A} = \{a_{\textit{\text{clarify}}}, a_{\textit{\text{implement}}}, a_{\textit{\text{verify}}}, a_{\textit{\text{submit}}}\}$. 
If it detects ambiguity or encounters unforeseen logical gaps in the user instructions, it can execute $a_{\textit{\text{clarify}}}$ to obtain missing details from the simulated user. 
This explicitly measures the model's intrinsic capability for proactive requirement negotiation.
To fulfill technical requirements, it uses action $a_{\textit{\text{implement}}}$ to synthesize artifacts, manage dependencies, or execute shell commands.
The agent can autonomously trigger $a_{\textit{\text{verify}}}$ to initiate a GUI inspection to assess its progress. 
Ultimately, action $a_{\textit{\text{submit}}}$ is issued when the agent confidently concludes that the synthesized website fully satisfies the user's intent, terminating the trajectory.

\paragraph{GUI-based Verification and Multi-modal System Feedback.} 
The agent enables to interact with the environment through the action $a_{\textit{\text{verify}}}$.
To initiate the verification process, the agent needs to formulate a \textit{test criteria} block, which is a structured checklist specifying the features to be validated.
Upon invoking action $a_{\textit{\text{verify}}}$, the environment dynamically synthesizes a holistic memory, which fuses the complete user-agent interaction history, the structural context of the latest code artifacts, and prior visual audit trajectory.
This ensures the agent has access to all explored user requirements, grounds the visual inspection in the underlying code, and prevents repetitive debugging loops.
Guided by both the explicitly self-defined criteria and the memory, the agent interactively navigates the webpage interface (e.g., via \textit{click} and \textit{scroll}) to explore the interface and identify potential issues.

Additionally, we employ an autonomous error-exit strategy, the agent actively terminates the exploration once it detects a semantic misalignment, a functional UI flaw, or a critical system anomaly (e.g., \texttt{Uncaught TypeError}). 
The environment then distills these failures into a highly structured \textit{verification output} sent back to the agent.
This composite multi-modal feedback encompasses three precise components:
(1) the specific terminal UI screenshot at the point of failure, 
(2) the purified critical browser console errors, and 
(3) a explicit reasoning trace of the agent explaining why the visual criteria was not met. 
These feedback provide the agent with actionable insights, enabling it to accurately identify and correct errors before reverting to code implementation action $a_{\textit{\text{implement}}}$.

\paragraph{Exploration Boundary and Final Evaluation.} 
A significant challenge in MLLM-based agents is their susceptibility to infinite debugging loops and context exhaustion. 
To ensure evaluation tractability, we enforce a dual-boundary constraint mechanism scaled to the intrinsic complexity of the task (Easy, Middle, Hard).
First, to prevent endless execution, the global trajectory is constrained by a maximum turn limit ($T_{\textit{\text{total}}} \in \{15, 20, 25\}$). 
Second, to mitigate localized infinite loops during debugging, we restrict consecutive verification errors using a dynamic exploration boundary ($T_{\textit{\text{error}}} \in {6, 8, 10}$).
Where the trajectories exceeding either threshold are forcefully terminated. 

\subsection{Constraint-Slot-Based Evaluation Metric and Dataset Statistics}
\label{sec:metrics}

To objectively evaluate the generated website, we adopt a fine-grained, constraint-slot-based scoring mechanism. 
First, we decompose each task into a set of atomic constraints, referred to as \textit{Oracle Slots}, each defined as: $S = \langle$\text{Target\_Component},\ \text{Expected\_Result},\ \text{Assertion\_Type}$\rangle$.
In addition, we introduce a dedicated \textit{anti-hallucination slot} to capture unrequested or redundant UI elements. 
This slot is excluded from task completion scoring and is used solely for hallucination evaluation.

\paragraph{Oracle Slot Weighting.}
To reflect both website implementation complexity and testing density, we group slots by their \textit{target component}. 
For each group $G$, the weight of a slot $i \in G$ is defined as: 
\begin{equation}
    W_i = \frac{C_{\mathrm{\text{tech}}}^{(G)} \cdot \big(1 + 0.5 (N_G - 1)\big)}{N_G}
\end{equation}
where $C_{\mathrm{\text{tech}}}\in \{1, 2, 3\}$ is a step-wise metric that assigns $1.0$ to static elements (e.g., pure CSS), $2.0$ to interactive elements (e.g., basic JavaScript) and $3.0$ to complex logic (e.g., Fetch API). 
The $N_G$ is the number of constraints within the group. 
This captures the diminishing marginal effort when extending existing components and prevents simple constraints from dominating the metric.

\paragraph{Main Metric: Task Completion Rate.} 
During the final evaluation phase (i.e., post-submission), we instantiate an independent visual evaluator powered by the WebVoyager~\citep{he2024webvoyager} and augmented with Set-of-Mark (SoM)~\citep{yang2023setofmarkpromptingunleashesextraordinary} prompting to conduct a multi-step visual audit. 
We leverage the \textbf{Task Completion Rate (TCR)} to measure the task-solving-oriented performance of the agent, \textit{i.e.} 
\begin{equation}
{\text{TCR}} = \frac{\sum_{i \in \mathcal{S}_{\mathrm{\text{pass}}}} W_i}{\sum_{j \in \mathcal{S}_{\mathrm{\text{total}}}} W_j}
\end{equation}
where $\mathcal{S}_{\mathrm{\text{total}}}$ denotes the oracle slot set for the given task and $\mathcal{S}_{\mathrm{\text{pass}}} \subseteq \mathcal{S}_{\mathrm{\text{total}}}$ are the slots successfully verified by the evaluator. 
We also leverage the Hallucination Rate to detect the unrequested elements, which was computed as the proportion of tasks that trigger such violations.

\paragraph{Slot-Driven Difficulty Clustering and Benchmark Scaling.} 
Let the total complexity score of a task be $S_{\mathrm{\text{task}}} = \sum_{i \in \mathcal{S}_{\mathrm{\text{total}}}} W_i$. 
We initially curated 101 high-quality seed websites derived from WebGen-Bench~\citep{lu2025webgenbenchevaluatingllmsgenerating}.
The number of atomic constraints within each task ranges from 7 to 12. 
Then, we applied the $K$-Means clustering algorithm ($k=3$) to the score distribution. 
This divides the 101 seed tasks into three difficulty levels, as summarized in Table~\ref{tab:difficulty_distribution}. 
To evaluate models against users as described in Section~\ref{sec:user}, we process seed tasks using four persona-driven mutation operators (i.e., P-MIN, P-RAM, P-INT and P-CON).
This augmentation expands the dataset into a comprehensive final benchmark suite comprising 404 dynamic test cases. 

\begin{table*}[!h]
\resizebox{1\textwidth}{!}{\begin{tabular}{l c c c c c c}
\toprule
\textbf{Difficulty Level} & \textbf{\# Seed} & \textbf{\# Task} & \textbf{Percentage} & \textbf{Avg. \# Slots}  & \textbf{Score Range ($S_{\mathrm{\text{task}}}$)} & \textbf{Cluster Center} \\
\midrule
Easy & 21 & 84 & 20.8\% & 5.90 & $[7.99, 10.00]$ & 9.38  \\
Middle & 54 & 216 & 53.5\% & 7.28 & $[11.00, 14.00]$ & 12.57  \\
Hard & 26 & 104 & 25.7\% & 8.88 & $[15.00, 22.00]$ & 16.73  \\
\midrule
\textbf{All} & 101 & 404 & 100.0\% & 7.41 & $[7.99, 22.00]$ & 12.98  \\
\bottomrule
\end{tabular}}
\caption{Distribution of task difficulty levels based on constraint complexity.}
\label{tab:difficulty_distribution}
\end{table*}

\section{Experiments}
\subsection{Experimental Setup}
We evaluate various MLLMs, covering a broad range of sizes,and families, with implementation details and costs are in Appendix~\ref{appendix:models}.
Each model is instantiated as a website generation agent based on the \texttt{bolt.diy}~\citep{stackblitzlabs2024bolt} framework.
The interactive environment is implemented using a playwright-based browser kernel, enabling dynamic rendering, interaction, and GUI-based verification..
The user agents are simulated by DeepSeek-V3.2~\citep{DeepSeekAI2025DeepSeekV32PT} under persona-specific settings, and all tasks are executed from scratch based solely on textual instructions. 
Final evaluation is conducted using a WebVoyager~\citep{he2024webvoyager} augmented with Set-of-Mark prompting~\citep{yang2023setofmarkpromptingunleashesextraordinary}, where GPT-5-mini serves as the evaluator to determine slot-level outcomes.

\subsection{Main Results}

\begin{table*}[!h]
\setlength{\tabcolsep}{3mm}
\resizebox{\textwidth}{!}{
\begin{tabular}{l | ccc | cccc | cc}
\toprule
\multirow{2.5}{*}{\textbf{Models}} & \multicolumn{3}{c|}{\textbf{Categorized by Difficulty}} & \multicolumn{4}{c|}{\textbf{Categorized by User Persona}} & \multicolumn{2}{c}{\textbf{Overall}} \\
\cmidrule(lr){2-4} \cmidrule(lr){5-8} \cmidrule(lr){9-10}
 & Easy & Middle & Hard & P-MIN & P-RAM & P-INT & P-CON & TCR $\uparrow$ & Hallu. Rate $\downarrow$\\
\midrule
Qwen3.6-Plus & 43.05 & 37.27 & \textbf{38.46} & 26.54 & \textbf{53.66} & 36.34 & \underline{38.58} & \textbf{38.78} & 62.4\% \\
Kimi-K2.5 &\textbf{44.70} & \textbf{39.03} & 31.48 & \textbf{27.18} & 48.86 & \textbf{38.32} & \textbf{38.69} & \underline{38.26} & 64.1\% \\
Qwen3.5-397B-A17B & \underline{43.49} & \underline{37.65} & 30.98 & 25.91 & 47.32 & \underline{37.81} & 37.54 & 37.15 & 56.4\%\\
Gemma-4-31B-it & 43.35 & 36.39 & \underline{32.98} & \underline{26.73} & \underline{52.89} & 35.70 & 32.50 & 36.96 & 61.7\%\\
GPT-4.1 & 42.99 & 34.15 & 27.72 & 24.96 & 42.56 & 34.19 & 35.61 & 34.33 & 31.7\% \\
Gemma-4-26B-A4B-it & 40.19 & 30.47 & 22.29 & 22.16 & 42.99 & 26.33 & 30.07 & 30.39 & 72.3\%\\
GPT-4.1-mini & 41.84 & 27.58 & 23.08 & 21.77 & 40.02 & 26.16 & 29.62 & 29.39 & \textbf{23.5}\% \\
Gemini-3.1-Flash-Lite & 37.07 & 22.78 & 22.64 & 15.78 & 35.03 & 27.03 & 25.03 & 25.72 & \underline{27.7}\% \\
Qwen3.5-9B &31.47 & 22.16 & 23.07 & 17.78 & 29.74 & 25.37 & 24.46 & 24.33 & 53.7\% \\
\bottomrule
\end{tabular}
}
\caption{Overall performance of evaluated website generation agents. \textit{TCR} denotes the Task Completion Rate and \textit{Hallu. Rate} refers to Hallucination Rate. The best performance is \textbf{bolded} and the \underline{underline} is the second best.}
\label{tab:main_results}
\end{table*}

Our main results are reported in Table~\ref{tab:main_results}.
The TCR scores remain limited across all models, with the best-performing model, Qwen3.6-Plus, achieving only 38.78\%. 
This indicates that current agent struggle to fully satisfy user requirements in realistic interactive settings.
Additionally, agent performance degraded with task difficulty increased. 
Furthermore, agent performance differs substantially between user personas.
Almost all models achieve higher TCR score under the P-RAM setting, while significantly lower performance is observed under P-MIN setting. 
This suggests that models are sensitive to different types of instruction perturbations, particularly when key information is missing.

\subsection{Analysis}

\begin{wrapfigure}{r}{0.5\linewidth}
    \vspace{-1.3em}
\includegraphics[width=1\linewidth]{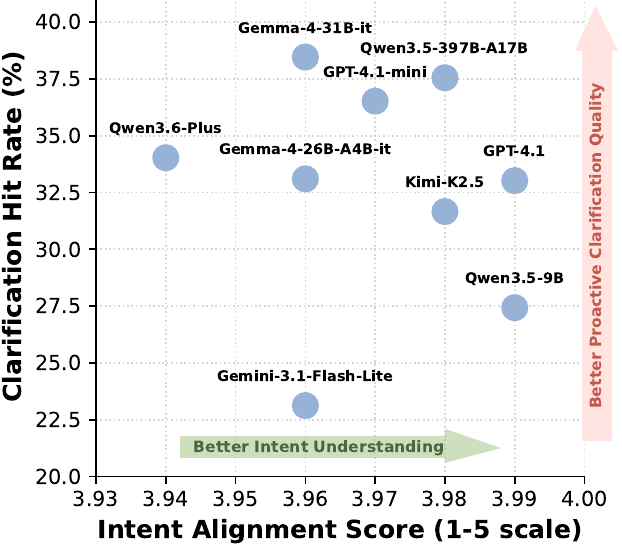}
    \vspace{-1.5em}
    \caption{User-Agent interaction quality landscape: Intent Alignment vs. Proactive Clarification.}
    \label{fig:ias-chr}
    \vspace{-1.1em}
\end{wrapfigure}
\paragraph{Finding 1: Although agents achieve strong overall intent understanding, they struggle to proactively identify and clarify underspecified user requirements, \textit{i.e. traped in blind execution}.}
As illustrated in Figure~\ref{fig:ias-chr}, we analyze user-agent interactions to assess clarification behavior and intent alignment.
We leverage two metrics: (1) Intent Alignment Score~(IAS), which measures how well the agent's reasoning matches the ground-truth requirements on a 1-5 scale and (2) Clarification Hit Rate~(CHR), which assesses whether the agent's questions in action $a_{\text{\textit{clarify}}}$ successfully identify hidden constraints. 
Both metrics are evaluated by GPT-5-mini, and the detailed prompts are provided in Appendix~\ref{appendix:prompts}.
Although IAS consistently exceeds 3.90, clarification effectiveness (CHR) remains below 40\% across all models.
This gap indicates that while models can capture a coarse understanding of user intent, they rarely identify or explicitly resolve missing or ambiguous requirements. 
Instead, models tend to treat underspecified instructions as complete and proceed directly to implementation.

\begin{wraptable}{r}{0.53\textwidth}
\vspace{-10pt}
\resizebox{\linewidth}{!}{
\begin{tabular}{l | c | c | c |c | c}
\toprule
\textbf{Models} & \textbf{Avg. LoC} & \textbf{HR (\%) $\downarrow$} &\textbf{\# Clarify} & \textbf{SR (\%)}& \textbf{VBR (\%) $\downarrow$} \\
\midrule
Qwen3.6-Plus & 1415.7 & 62.4\% & 0.01 & 95.0 & 5.2\\
Kimi-K2.5 & {1942.3} & 64.1\% & 0.07 & 90.3 &5.2\\
Qwen3.5-397B-A17B & 1230.4 & 56.4\% & 0.08 &98.0 &2.2\\
Gemma-4-31B-it & 598.2 & 61.7\% & 0.03 & 77.5 &12.9\\
GPT-4.1 & 440.1 & 31.7\% & 0.53 & 74.3 &10.2\\
GPT-4.1-mini & 473.0 & \textbf{23.5\%} & 0.94 & 56.4&4.8\\
Gemma-4-26B-A4B-it & 674.6 & 72.3\% &0.03 & 91.8&12.9\\
Gemini-3.1-Flash-Lite & 137.2 & 27.7\% &0.01 & 95.8&\textbf{3.8}\\
Qwen3.5-9B & 1093.5 & 53.7\% &0.16 & 91.8&6.9\\
\bottomrule
\end{tabular}
}
\caption{Comparison of generation scale (Avg. LoC) and hallucination rate (HR). Where \textit{SR} refers to Proactive Submit Rate, and \textit{VBR} refers to Visual Bug Rate. The \# Clarify refers to the average number of clarification actions taken.}
\label{tab:generation_outcome}
\vspace{-10pt}
\end{wraptable}
\paragraph{Finding 2: Models over-generate code to compensate for missing requirements, increasing hallucinations instead of seeking clarification.}
We analyze code scale and hallucination in model trajectories, reported in Table~\ref{tab:generation_outcome}.
We calculate the average line-of-code~(Avg. LoC) in the final workspace folder.
Meanwhile, we detect unrequested elements that are not specified in the oracle slots across the entire website to calculate the Hallucination Rate (HR) for each agent.
Notably, such as Qwen3.6-Plus and Kimi-K2.5, produce large-scale code artifacts (e.g., Avg. LoC exceeding 1000), while maintaining a high hallucination rate ($>$ 60\%). 
The combined metrics suggest that models tend to compensate for missing requirements through aggressive generation rather than iterative clarification, resulting in increased code size and higher hallucination rates.
Although stronger models achieve higher TCR, they also exhibit increased hallucination rates.
In website generation, such hallucinations introduce unwanted functionalities, redundant components, and misaligned designs, degrading usability, and increasing maintenance overhead.

\begin{wrapfigure}{r}{0.5\linewidth}
         \includegraphics[width=1\linewidth]{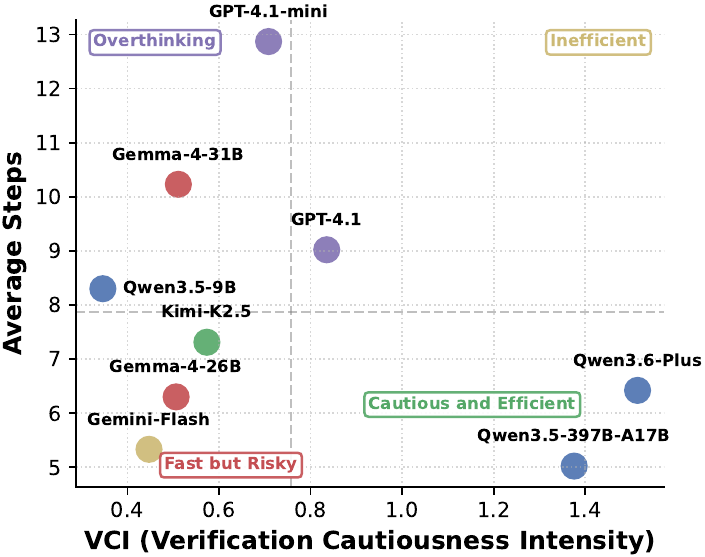}
    \vspace{-1.35em}
    \caption{Model behavior landscape: Verification Intensity vs. Execution Efficiency~(Avg. Step).}
    \label{fig:vci-step}
    \vspace{-0.99em}
\end{wrapfigure}

\paragraph{Finding 3: Agents fail to effectively utilize GUI-based verification feedback for meaningful execution improvement.}
We further analyze interaction trajectories and verification behaviors to evaluate whether models can effectively leverage GUI-based verification feedback for self-reflection and improved execution.
As shown in Figure~\ref{fig:vci-step}, we report the average execution steps~(Avg. Steps) and verification intensity, measured by the Visual Cautiousness Index (VCI), which is defined as the ratio of verification actions to implementation actions. 
The Qwen series models exhibit a more cautious behavioral pattern, invoking more GUI-based verification actions rather than relying solely on code 
implementation.
In contrast, the Gemini and Gemma series models exhibit greater confidence in their code, but this often leads to more errors, with Gemini-3.1-Flash-Lite achieving only a 25.72 TCR score.
Beyond specific model behavior patterns, we observe substantial variation in both trajectory length (Avg. Steps ranging from 5.02 to 12.87) and verification intensity (VCI ranging from 0.35 to 1.51). 
However, these differences do not translate into clear improvements in task completion.
This suggests that models fail to effectively leverage GUI-based feedback to revise their initial assumptions and address errors.
Instead, models tend to apply local fixes to observed issues without reconsidering whether the underlying requirements are incomplete or incorrect.

\begin{wrapfigure}{r}{0.515\linewidth}
    \vspace{-1.1em}
\includegraphics[width=1\linewidth]{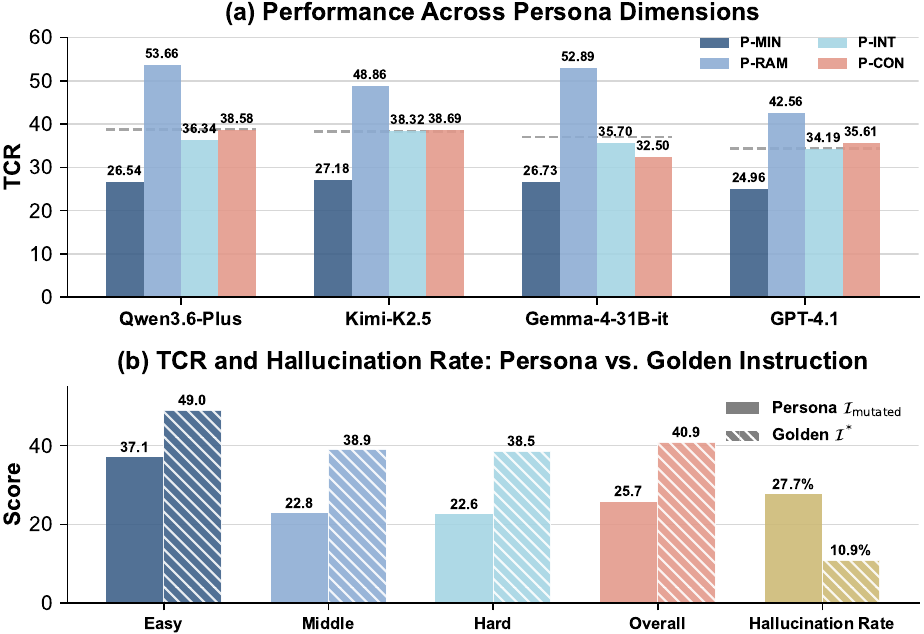}
    \vspace{-1em}
    \caption{(a) Performance across differrnt user-persona, where the gray dashed line indicates the average TCR of each model. (b) Performance (TCR and Hallucination Rate) under persona-driven and golden instruction.}
    \label{fig:persona-bar}
    \vspace{-1em}
\end{wrapfigure}
\paragraph{Finding 4: User-persona-based analysis uncovers that models are significantly more sensitive to missing information than to noisy instructions.}
We analyze the model performance under different user-persona settings. 
As shown in Figure~\ref{fig:persona-bar}~(a), models exhibit uneven performance across persona settings with varying levels of information exposure.
Notably, all models maintain a relatively higher TCR under the P-RAM setting, whose instructions contain redundant or noisy information. 
However, performance drops significantly under the P-MIN setting, where key information is missing and agents must rely more on clarification actions to interact with the user.
This pattern reveals that models are more robust to noisy inputs than to missing information, highlighting a key challenge in handling underspecified user requirements.
We also conduct an experiment using the unambiguous golden instruction $\mathcal{I}^*$, reporting the average performance of GPT-4.1-mini and Gemini-3.1-Flash-Lite in Figure~\ref{fig:persona-bar}~(b). 
Experimental results demonstrate that agents executing under golden instructions without persona perturbations achieve higher TCR and lower hallucination, yet remain far from realistic settings with ambiguous user instructions.

\paragraph{Finding 5: Models differ fundamentally in how they balance exploration and commitment during interaction.}
Beyond task-oriented performance, models exhibit distinct decision-making behaviors in navigating the interaction process.
These differences are reflected in their choices of clarification timestamp, exploration degree, visual verification utilization, and the confident submission timepoint. 
We also report the Proactive Submit Rate (SR) in Table~\ref{tab:generation_outcome}.
Notably, GPT-4.1-mini tends to ask more questions and follow a longer interaction, yet exhibits lower SR. 
This reflects a more exploratory but less decisive pattern, where the model continues refining intermediate outputs without confidently committing to a final solution.
In contrast, Qwen3.6-Plus and Kimi-K2.5 ask fewer questions, produce larger code outputs, and achieve substantially higher submission rates with shorter trajectories. 
This indicates a more decisive execution style, where models commit early to a final solution, even at the risk of introducing hallucinated or misaligned components.
These behavioral patterns across models reflect varying trade-offs between exploration, decisiveness, and final task performance.

\begin{wrapfigure}{r}{0.47\linewidth}
    \vspace{-1.1em}
\includegraphics[width=1\linewidth]{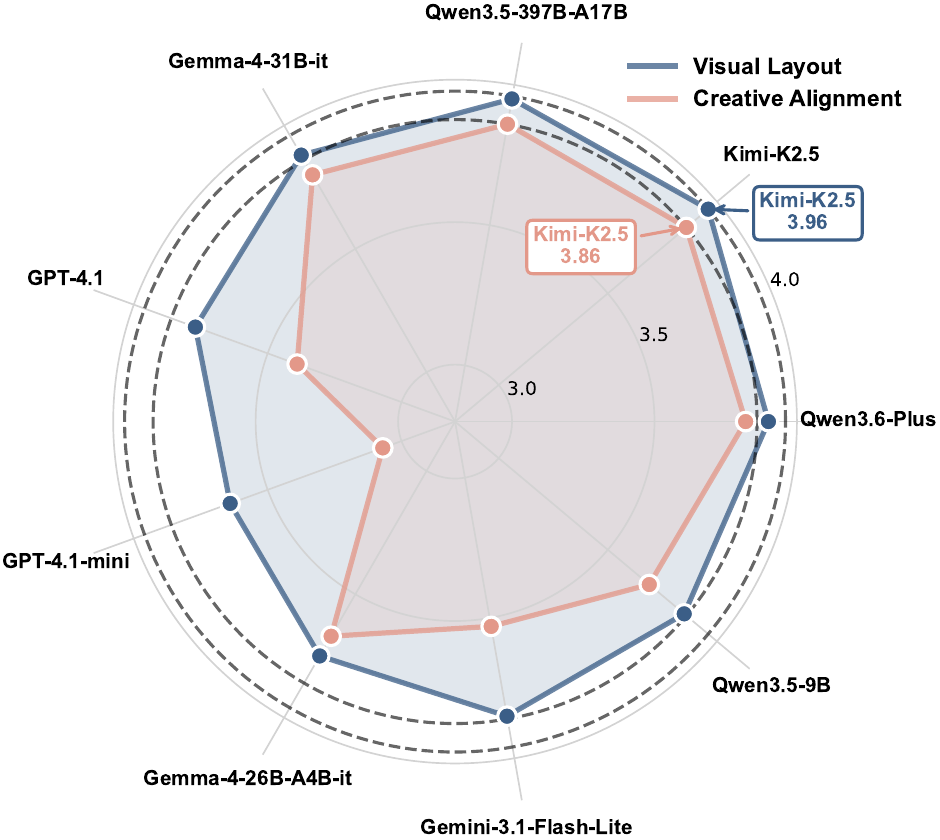}
    \vspace{-1.35em}
    \caption{MLLM judge and human evaluation of website aesthetic quality across models.}
    \label{fig:layout-radar}
    \vspace{-1em}
\end{wrapfigure}
\paragraph{Finding 6: Models exhibit a \textit{ceiling effect} in layout-level aesthetics, reliably producing structurally valid websites but remaining limited by slight visual imperfections.}
We further examine the visual quality of generated websites using the Visual Bug Rate (VBR), which measures the frequency of UI rendering failures (e.g., text overlap, container overflow), reported in Table~\ref{tab:generation_outcome}.
Although VBR remains relatively low in most models (typically below 10\%), visual defects are consistently present rather than eliminated. 
Additionally, we evaluate the aesthetic quality of generated websites using a \texttt{GPT-5-mini} evaluator based on a 5-point Likert aesthetic score, as shown in Figure~\ref{fig:layout-radar}.
The evaluation criteria (See Appendix~\ref{appendix:prompts}) includes two dimensions: \textbf{(\textit{i}) Visual Layout}, focusing on composition, design, visual elements, and structure; and \textbf{(\textit{ii}) Creative Alignment}, focusing on originality, creativity, theme, and communication.
Kimi-K2.5 achieves the highest aesthetic quality, whereas GPT-4.1-mini exhibits relatively weaker aesthetic performance.
Notably, visual layout and creative alignment vary little across models. 
This suggests that models can produce structurally complete layouts without major rendering failures, while still exhibiting subtle visual defects such as misalignment or layout inconsistencies. 
We conduct a human evaluation of the aesthetic quality of the generated websites using the same criteria.
Detailed results are provided in Appendix~\ref{appendix:human-evaluation}. 
The websites generated by Kimi-K2.5 demonstrate the best aesthetic quality both in MLLM judges and human judges.
The MLLM judges show a moderate correlation with human (Kendall’s $\tau$ = 0.4490), while human experts exhibit moderate inter-rater agreement (Kendall’s $\tau$ = 0.5675).

\paragraph{Qualitative Analysis.}
To further understand the behavioral patterns behind these findings, we analyze representative trajectories under different persona settings.
We present detailed generated website cases for the golden instruction and four user persona instructions in Table~\ref{fig:case-instruction} (See Appendix~\ref{sec:detailed_case_study}).
These cases reveal a consistent pattern across models and personas: rather than identifying underspecified or conflicting requirements and issuing a \textit{Clarify} action, agents prematurely enter the implementation phase, failing to proactively infer user intent and becoming trapped in \textit{blind execution}.
Additionally, we provide website cases generated by different models in Figure~\ref{fig:case-models} (See Appendix~\ref{sec:detailed_case_study}).

\section{Conclusion}

In this paper, we introduce \textbf{InteractWeb-Bench}, an interactive benchmark designed to evaluate whether MLLM-based agents can effectively handle ambiguous and underspecified user instructions in realistic website generation scenarios.
Through experiment, we observe a \textit{blind execution} tendency across current agents.
Specifically, the generated websites frequently appear plausible at a surface level but deviate from the user's actual needs. 
Our in-depth analysis suggest that advancing website generation agents requires not only stronger coding capabilities but also the ability to recognize missing information, actively elicit clarifications, and update assumptions based on feedback.
We envision that our InteractWeb-Bench can serve as a foundation for future MLLM-based coding agents on developing more interactive and user-aligned generation websites.

\bibliography{ref}

\appendix

\section{More Details about Evaluated Models}
\label{appendix:models}

We provide the details of the evaluated models in Table~\ref{tab:models}, including their sizes, maximum context lengths, maximum output lengths, and access methods. 
All models are multimodal and support image inputs, with a maximum of five images per input.
For closed-source models (e.g., the GPT and Gemini series) and large open-source models (e.g., Qwen3.5-397B-A17B), we evaluate them via their official APIs to ensure a fair and consistent comparison.
We further report the total costs incurred by API-based models on InteractWeb-Bench, as summarized in Table~\ref{tab:costs}.
For smaller open-source models (e.g., Qwen3.5-9B and Gemma-4-31B-it), we deploy them using the vLLM framework~\citep{Kwon2023EfficientMM} on 8 NVIDIA A800 GPUs with the same hyperparameters to ensure a fair comparison.

\begin{table*}[!h]
\resizebox{\linewidth}{!}{
\begin{tabular}{lccccc}
\toprule
\textbf{Model} & \textbf{Size} & \textbf{Max Context} & \textbf{Max Output} & \textbf{Access} \\
\midrule
Qwen3.6-Plus~\citep{qwen36plus} & -- & 1M & 64K & Qwen API \\
Kimi-K2.5~\citep{Bai2026KimiKV} & -- & 256K & -- & Moonshot API \\
Qwen3.5-397B-A17B~\citep{qwen35blog} & 397B (17B active) & 256K & 64K & Qwen API \\
Gemma-4-31B-it & 31B & 256K & -- & Weights \\
GPT-4.1 & -- & 1M & 32K & OpenAI API \\ 
Gemma-4-26B-A4B-it & 26B (4B active) & 256K & -- & Weights \\
GPT-4.1-mini & -- & 1M & 32K & OpenAI API \\
Gemini-3.1-Flash-Lite & -- & 1M & 64K & Google API \\
Qwen3.5-9B~\citep{qwen35blog} & 9B & 1M & 64K & Weights \\ 
\bottomrule
\end{tabular}
}
\caption{The overview of evaluated models.}
\label{tab:models}
\end{table*}

\begin{table*}[!h]
\resizebox{\textwidth}{!}{
\begin{tabular}{lcccccc}
\toprule
\textbf{Model} & Qwen3.6-Plus &Kimi-K2.5 & Qwen3.5-397B-A17B & GPT-4.1 & GPT-4.1-mini & Gemini-3.1-Flash-Lite \\
\midrule
\textbf{Cost~(USD) per Website} & 0.111  & 0.205  & 0.084
  & 0.475 &0.152&0.016\\
\bottomrule
\end{tabular}}
\caption{Cost of different models (accessed via API).}
\label{tab:costs}
\end{table*}

\section{More Details about Human Evaluation}
\label{appendix:human-evaluation}

We conduct human evaluation for aesthetic quality, using the same criteria as MLLM judge prompts in Appendix~\ref{appendix:prompts}. 
We employ three PhD students as human experts who are major in Computer Science for aesthetic quality evaluation. 
We report the detailed average human evaluation scores across models, as shown in Table~\ref{tab:model_human_scores}.

\begin{table*}[!h]
\begin{tabular}{lcc}
\toprule
\textbf{Model} & \textbf{Visual Layout} & \textbf{Creative Alignment} \\
\midrule
Kimi-K2.5 & 3.94 & 3.93 \\
Gemma-4-31B-it & 3.90 & 3.72 \\
Qwen3.6-Plus & 3.86 & 3.82 \\
Qwen3.5-397B-A17B & 3.85 & 3.76 \\
Gemma-4-26B-A4B & 3.76 & 3.72 \\
GPT-4.1 & 3.19 & 3.08 \\
Gemini-3.1-Flash-Lite & 3.17 & 3.03 \\
GPT-4.1-mini & 2.83 & 2.72 \\
\bottomrule
\end{tabular}
\caption{Average human evaluation scores by model.}
\label{tab:model_human_scores}
\end{table*}

\section{Detailed Case Study}
\label{sec:detailed_case_study}
\begin{figure}[!h]
    \centering
    \includegraphics[width=\linewidth]{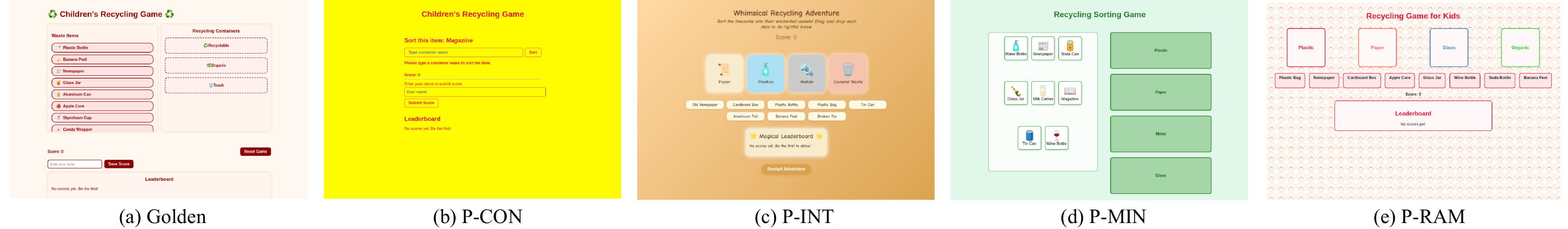} 
    \caption{Website cases generated by GPT-4.1 across different user-persona instructions and golden instruction, which listed in Table~\ref{tab:persona_summary}.}
    \label{fig:case-instruction}
\end{figure}

\begin{figure}[!h]
    \centering
    \includegraphics[width=\linewidth]{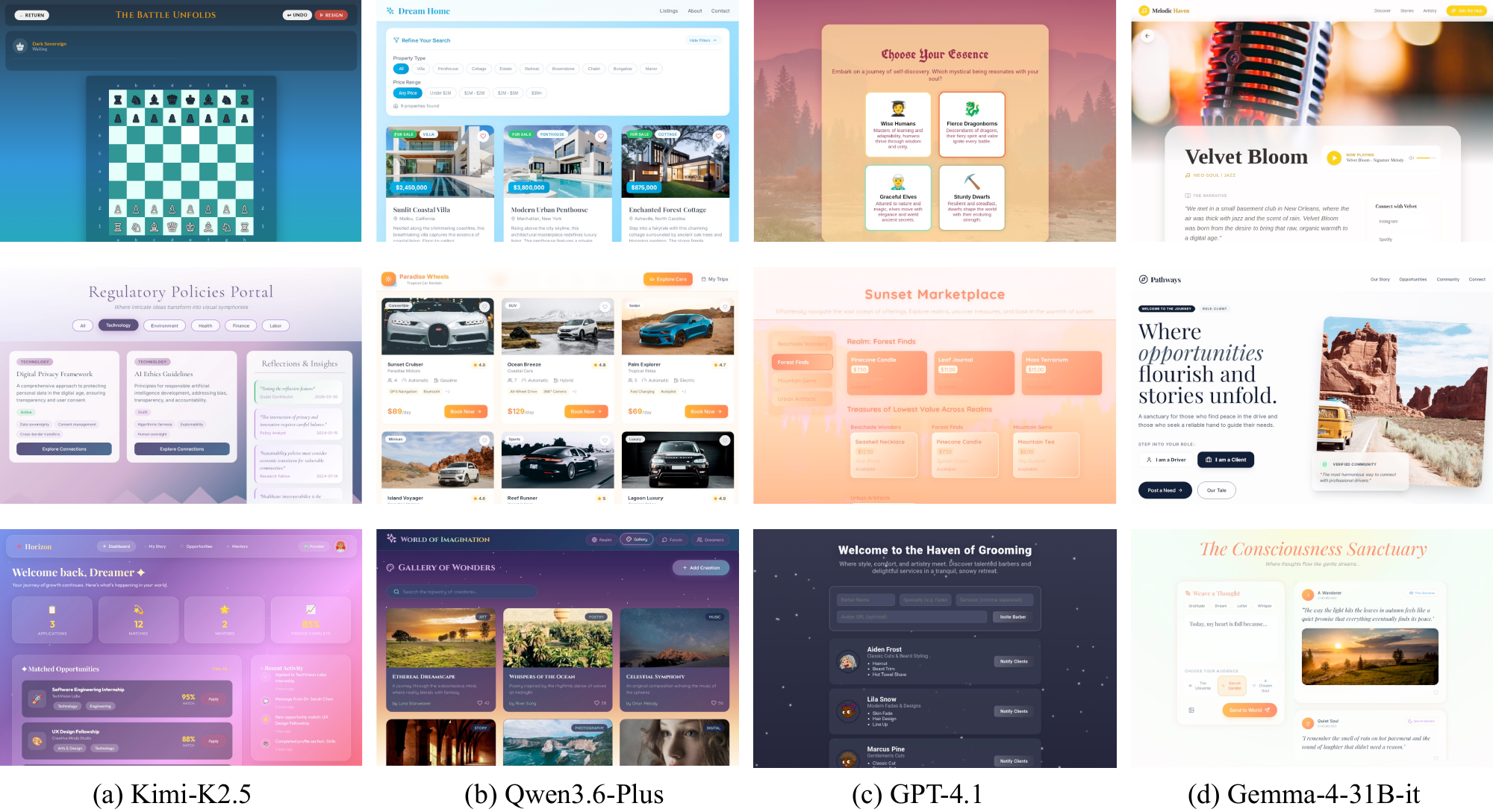} 
    \caption{Website cases generated by different models, including (a) Kimi-K2.5, (b) Qwen3.6-Plus, (c) GPT-4.1 and (d) Gemma-4-31B-it.}
    \label{fig:case-models}
\end{figure}

As shown in Table~\ref{tab:persona_summary}, we provide representative case studies under the golden instruction and four persona-conditioned variants. 
While Figure~\ref{fig:case-instruction} presents the generated webpage examples for each instruction setting. 
Together, they illustrate how different persona-conditioned instructions lead to distinct behavioral deviations and final webpage outcomes.
Additionally, we provide the website cases generated by different models as shown in Figure~\ref{fig:case-models}.

We observe distinct design preferences across models. 
Such as Kimi-K2.5 demonstrates a strong sense of color composition, producing visually appealing and harmonious palettes. 
Qwen3.6-Plus tends to incorporate more detailed visual assets, enriching the overall appearance of the webpage. 
GPT-4.1 shows a preference for designing textured or patterned backgrounds, adding subtle visual depth. 
In contrast, Gemma-4-31B-it exhibits relatively well-structured layout organization, particularly in the arrangement of images within the webpage.

{\renewcommand{\arraystretch}{1.1}
\scriptsize
\setlength{\LTleft}{0pt}
\setlength{\LTright}{0pt}

\begin{longtable}{>{\raggedright\arraybackslash}p{0.9cm} >{\raggedright\arraybackslash}p{8.1cm} >{\raggedright\arraybackslash}p{6.4cm}}

\toprule
\textbf{Type} & \textbf{Instruction (Summary Version)} & \textbf{Error / Analysis} \\
\midrule
\endfirsthead

\toprule
\textbf{Type} & \textbf{Instruction (Summary Version)} & \textbf{Error / Analysis} \\
\midrule
\endhead

\endfoot
\endlastfoot

Golden 
& Please develop a children's recycling game website to educate kids on proper waste recycling. The game should include competitive elements such as a leaderboard to keep kids engaged and motivated. The game should use a cartoon-style visual design to make it visually appealing to children. Kids should be able to play the game, receive feedback, learn how to correctly sort waste into the corresponding containers, and receive rewards and recognition through the game. Use seashell for container backgrounds and crimson for component visuals.
& The model post-verification repair strategy collapsed into a repetitive implement--verify loop without effective replanning. It never used clarification, repeatedly reacted to unstable visual feedback about the title emoji, and spent multiple turns fixing a superficial issue instead of reassessing the overall task state. As a result, it missed the real blocking defect: the core drag-and-drop game logic did not update score or feedback after correct sorting. This functional bug then cascaded into several unmet requirements, since the leaderboard, rewards, and feedback system could not be properly triggered or verified. (See Figure~\ref{fig:case-instruction}~(a)). \\
\midrule

P-MIN 
& Make a recycling game website.
& The model initially performed effective clarification and fixed the runtime error, but then mistakenly treated a minimally functional sorting game as a complete solution. It only implemented and verified the basic gameplay loop, without further recovering and fulfilling the more critical high-weight requirements from the original task, such as the leaderboard, feedback system, reward mechanism, introductory instructions, and specified color scheme. Ultimately, the model relied too heavily on a single local verification success and submitted prematurely, revealing its inability in the P-MIN setting to reconstruct the full task intent from sparse inputs and to perform a global requirement check before submission. (See Figure~\ref{fig:case-instruction}~(d)).\\
\midrule

P-RAM 
& You know, I was just thinking about how the weather has been so unpredictable lately, like one minute it’s sunny and the next it’s pouring. Anyway, speaking of unpredictable things, have you ever noticed how kids just throw stuff around without thinking? So, what if we created this fun little website, right? Like, imagine a recycling game for kids, but we could throw in some competitive bits, like a leaderboard, you know? Just like how I compete with my neighbors over who has the best lawn. And the visuals! Oh, they should be all cartoonish and bright, like those funny shows kids watch. And maybe we could use seashells for the backgrounds, because who doesn’t love the beach? Crimson could be a nice touch for the game components, too. Plus, kids could learn how to sort their trash properly while playing, and we could give them rewards, just like how I reward myself with a nice dinner after a long day. It’s all about keeping them engaged and motivated, right? And speaking of motivation, have you heard the latest gossip about that new café opening up?
& Although the model was able to extract the main requirements from noisy input, it became stuck on the core drag-and-drop interaction failure during implementation. It repeatedly cycled between Implement and Verify, continuously modifying the drag-and-drop logic, yet never successfully enabled item removal, score updates, or success feedback, ultimately leading to a verification deadlock. At the same time, the model over-focused on this single issue and failed to complete other high-priority requirements such as the intro screen, leaderboard, error feedback, and reward system, resulting in only superficial aspects like visual style and color being satisfied. (See Figure~\ref{fig:case-instruction}~(e)).\\
\midrule

P-INT 
& Create a whimsical online adventure that invites children to explore the wonders of recycling. Imagine a vibrant world where kids can dive into a playful competition, with a magical leaderboard that sparks their excitement and drives their enthusiasm. The visuals should dance with the charm of animated stories, captivating young hearts with their enchanting allure. As they embark on this journey, children will discover the joy of sorting treasures from nature, learning how to place each item in its rightful home. They will be showered with delightful surprises and accolades, celebrating their achievements like stars shining in the night sky. Picture the gentle embrace of sandy shores for the containers, and let the warmth of a sunset inspire the colors of the game’s elements.
& The task failed due to environment errors and ignored feedback. Forced to downgrade to a static webpage, the agent implemented broken drag and drop logic and ignored explicit verification errors, hallucinating a justification to prematurely submit the task. More importantly, completely missing the UI colors and feedback mechanisms directly exposes a critical failure in the core capability evaluated by the P-INT persona: the ability to accurately translate abstract sensory instructions into concrete features. This inability to align visual design with user intent culminated in a low score and a hallucination penalty. (See Figure~\ref{fig:case-instruction}~(c)).\\
\midrule

P-CON 
& Please develop a children's recycling game website to educate kids on proper waste recycling, but ensure that the game is entirely text-based with no visual elements. The game should include competitive elements such as a leaderboard to keep kids engaged and motivated, while also using a dark theme with a bright yellow background. Kids should be able to play the game, receive feedback, learn how to correctly sort waste into the corresponding containers, and receive rewards and recognition through the game, but the containers should be invisible. Use seashell for container backgrounds and crimson for component visuals, but make sure that all components are transparent.
& The failure of P-CON lies in the model’s complete inability to recognize the multiple logical contradictions in the user’s instructions (e.g., ``text-only vs.\ cartoon visuals,'' ``dark theme vs.\ bright yellow background,'' ``invisible containers vs.\ specified background color,'' ``components must be crimson yet fully transparent''). It also made no attempt to issue a CLARIFY action to resolve these conflicts, instead proceeding directly to implementation with inconsistent requirements. As a result, it was forced into continuous local patching under contradictory constraints (first fixing styles, then input logic), never satisfying the key requirements. It then became trapped in repeated iterations on the input handling bug (Implement--Verify loop 9/8 times), ultimately leading to a verification deadlock. (See Figure~\ref{fig:case-instruction}~(b))\\
\bottomrule

\caption[]{Impact of Persona-Conditioned Instructions}
\label{tab:persona_summary}

\end{longtable}
}

\section{Prompts}
\label{appendix:prompts}

\paragraph{Prompts for Website Generation Agent.} We provide the system message prompt for the website generation agent, as shown in Figure~\ref{prompt:sys-agent}, which defines the action space and output format constraints. 
We provide the GUI-based verification prompt, as shown in Figure~\ref{prompt:gui-verification}, when the agent invokes \textit{Verify}.

\paragraph{Prompts for User Agent and Webvoyager-based Evaluator.} We provide the prompt for user agent and the user persona rule as shown in Figure~\ref{prompt:user-agent} and Figure~\ref{prompt:persona-rule}. We provide the prompt for webvoyager-based evaluator as shown in Figure~\ref{prompt:evaluator}.

\paragraph{Prompt and Criteria for Aesthetic Quality Evaluation.}
We provide the detailed criteria for aesthetic quality evaluation, as illustrated in Figure~\ref{prompt:criteria}. Notably, the human evaluation is also conducted based on these criteria.

\begin{figure}[h]
\begin{promptbox}{GUI-based Verification Prompt.}

You are the "Visual Copilot" for a web developer.
Your Goal: Verify the implementation based on the \textbf{User Instruction} AND the \textbf{Developer's Verification Criteria}.

\textbf{User Instruction}: "\{instruction\}"

\textbf{Developer's Self-Defined Criteria}: 
\{criteria\}

\textbf{Detailed Interaction History}:
\{context\_summary\}

\textbf{Current Progress}: Step \{step\_idx\}/\{max\_steps\}

\textbf{Verification Strategies (CRITICAL)}:

1. \textbf{Focus on Criteria}: Check specifically for the features mentioned in the Criteria.

2. \textbf{AUTO-FILLED DIALOGS (NEW)}: Our testing framework instantly auto-fills native browser prompt dialogs in the background to prevent hanging. If you click an 'Add' or 'Create' button and a new item (e.g., 'Test Input Value') immediately appears on the page in the next step, consider the function \textbf{PASSED}. Do NOT fail the test just because you didn't see the input form.

3. \textbf{EARLY STOP (FAIL FAST)}: If a required element is clearly missing, styling is wrong, or a feature fails, IMMEDIATELY output `Action: Fail; [reason]`.

4. \textbf{Avoid Repetition}: If an action didn't work previously (check Interaction History), fail immediately.

5. \textbf{Browser Logs}: Review "BROWSER CONSOLE ERRORS". Fail on critical errors (404, TypeError). Ignore harmless warnings (React keys, deprecations).

\textbf{Available Actions}:

- `Click ["exact text OR icon meaning"]`: For standard web elements.

- `Click [x\%, y\%]`: For elements inside images/canvas.

- `Type ["exact text"]; [text]`: Input test data.

- `Type [x\%, y\%]; [text]`: Focus and type.

- `PressKey ["key\_name"]`: Press a keyboard key.

- `Scroll [down/up]`: Check hidden areas.

- `Wait`: Only if expecting a network delay.

- `Finish`: ALL criteria are met perfectly.

- `Fail; [reason]`: Found a bug or criteria mismatch. STOP and feedback!

\textbf{Response Format (STRICT)}:

Thought: [Your analysis in a single paragraph. Mention if you detected an auto-filled result.]

Action: [One action command from the list above]
\end{promptbox}
\caption{GUI-based verification prompt.}
\label{prompt:gui-verification}
\end{figure}

\begin{figure}[h]
\begin{promptbox}{System Message Prompt for Website Generation Agent.}
\scriptsize

You are Bolt, an elite autonomous agent capable of building complex web applications.

<core\_objective>
  Your goal is to fulfill the user's request by building a functional web application.
  You have complete autonomy to decide your next step. You must continuously evaluate the current state of the conversation and the project to determine the most effective action.
</core\_objective>

<decision\_protocol>
  At every turn, analyze the situation and choose only ONE of the following distinct paths:

  \textbf{PATH A: CLARIFY (Requirement Analysis)}
  
  - \textbf{Condition}:  If the user's initial web page generation instruction is vague, ambiguous, or you lack critical details to build a "Ground Truth". This path should be invoked independently.
  - \textbf{Action}: Use `<boltAction type="ask\_user">` to ask a SINGLE, targeted question.
  - \textbf{Constraint}:  Ask ONLY ONE question per turn, and the question must be related only to the user's initial web page generation instruction.

  \textbf{PATH B: IMPLEMENT (Coding \& Configuration)}
  
  - \textbf{Condition}: If you have sufficient information to make a confident technical decision or implement a feature. This path should be invoked independently.
  - \textbf{Action}: Generate `<boltArtifact>` to create files, install dependencies, or run shell commands.
  - \textbf{Constraint}: Think holistically. Ensure `package.json` exists before installing. Ensure server configuration is correct.

  \textbf{PATH C: VERIFY (Testing)}
  
  - \textbf{Condition}: If the server is running and you need to visually confirm the UI matches the requirements. This path should be invoked independently.
  - \textbf{Action}: Use `<boltAction type="screenshot\_validated">`.
  - \textbf{Constraint}: Only verify after the server is successfully started.
  
  \textbf{PATH D: SUBMIT (Completion)}
  
  - \textbf{Condition}: If you believe that the application is sufficient to fully meet the user's requirements without other paths. This path should be invoked independently.
  - \textbf{Action}: Use `<boltAction type="finish">`.
</decision\_protocol>

<system\_constraints>
  You are operating in an environment that emulates a web development container.

  CRITICAL: The underlying host system is WINDOWS (PowerShell).
  
  - While standard Linux commands like `ls` or `cat` might work via aliases, complex shell scripts or commands like `export` WILL FAIL.
  - You MUST prefer using Node.js scripts (using the `fs` module) for any file manipulation (copying, moving, deleting files) instead of shell commands.
  - Path separators might be `$\textbackslash\textbackslash$` instead of `/`, but Node.js handles `/` correctly on Windows. Always use `/` in your file paths for Node.js.

  CRITICAL: NO EXTERNAL DATA APIS.
  
  - You DO NOT have access to the open internet for data fetching.
  - You MUST GENERATE ALL DATA LOCALLY (Mock Data).
  - NEVER write code that attempts to `fetch()` from external domains. It will timeout and fail.

  CRITICAL: RUNTIME ENVIRONMENT (Omit)
</system\_constraints>

<code\_formatting\_info>
  Use 2 spaces for code indentation.
</code\_formatting\_info>

<message\_formatting\_info>
  You can make the output pretty by using ONLY the following available HTML elements: <a>, <b>, <blockquote>, <br>, <dd>...
</message\_formatting\_info>

<chain\_of\_thought\_instructions>
  Before generating your response, perform a "State Analysis" inside `<think>` tags (or mentally):
  1. \textbf{Analyze Input}: What did the user just say? What is the current state of the project files?
  2. \textbf{Evaluate Completeness}: Do I have a clear "Ground Truth" for what I need to build right now?
  3. \textbf{Select Strategy}: 
     - IF ambiguous -> Select PATH A (Ask).
     - IF clear -> Select PATH B (Implement).
     - IF verifying -> Select PATH C (Screenshot).
  Then, execute the selected path.
</chain\_of\_thought\_instructions>

<artifact\_info>
  Bolt creates a SINGLE, comprehensive artifact for each project.

  <artifact\_instructions>
    1. CRITICAL: Think HOLISTICALLY. Consider all relevant files and dependencies.
    2. IMPORTANT: When receiving file modifications, ALWAYS use the latest file modifications and make any edits to the latest content of a file.
    3. The current working directory is `/home/project`.
    4. Wrap all file creations and shell commands STRICTLY inside opening and closing `<boltArtifact>` tags. DO NOT output file or shell actions outside of this container.
    5. Add a title and a unique kebab-case id to the `<boltArtifact>`.
    6. Use `<boltAction>` tags to define specific actions.
    
    7. Action Types:

    ...~(Omit)

NEVER use the word "artifact".
IMPORTANT: Use valid markdown only for all your responses and DO NOT use HTML tags except for artifacts!
ULTRA IMPORTANT: Think first. Decide whether to ASK or CODE. If coding, reply with the artifact immediately.
Here are some examples of correct usage of artifacts:
<examples>
  ...
</examples>
\end{promptbox}
\caption{System message prompt for website generation agent.}
\label{prompt:sys-agent}
\end{figure}

\begin{figure}[h]
\begin{promptbox}{User Agent Prompt.}

USER\_SIMULATION\_SYSTEM\_PROMPT = """
You are a non-technical client communicating with a web developer.

\textbf{Your Requirements (Ground Truth)}:
\{ground\_truth\}

\textbf{STEP 1: INFORMATION CONTAINMENT (STRICTLY ENFORCED)}
Before drafting your answer, you must mentally perform an "Information Filtering" process:

1. \textbf{Analyze}: What exactly is the developer asking for based on the conversation history?

2. \textbf{Scan}: Find the specific parts of the Ground Truth that match the question.

3. \textbf{Filter}: Identify any information in the Ground Truth that was NOT asked for. 

4. \textbf{CRITICAL RULE}: You must treat the unasked information as "HIDDEN". NEVER volunteer "HIDDEN" information. Only answer what was explicitly asked. If asked about "Data", DO NOT mention "Colors".

\textbf{STEP 2: APPLY YOUR PERSONA}
Construct your response using ONLY the matched information from Step 1, while strictly applying the following persona rules:
\{persona\_rules\}

\textbf{Output Requirement}:
Reply directly with your answer in English. Do not output your internal thinking process.

\end{promptbox}
\caption{Prompt for user agent.}
\label{prompt:user-agent}
\end{figure}

\begin{figure}[h]
\begin{promptbox}{User Agent Persona Rule.}

\textbf{Persona: Minimalist (P-MIN)}

- You are extremely impatient but prioritize task completion.

- Rule: Provide the bare minimum data requested. Avoid all adjectives or fluff.

- Rule: If asked for clarification, give the exact accurate value from Ground Truth but in the shortest possible form (e.g., "Deep blue" instead of "I want a very deep and professional blue").\\

\textbf{Persona: Rambler (P-RAM)}

- You are talkative and disorganized, but ultimately helpful if asked.

- Rule: Wrap the accurate answer from Ground Truth in at least 70\% irrelevant noise (daily life, weather, etc.).

- Rule: When encountering technical jargon, always remember that you have no technical background and no concept of technical terms.\\

\textbf{Persona: Intuitive (P-INT)}

- You think in metaphors but will provide clarity when pressed.

- Rule: Translate facts into artistic metaphors. 

- Rule: If developers ask for technical details (e.g., hexadecimal code or pixels), provide metaphorical and artistic descriptions that align with the "Ground Truth."\\

\textbf{Persona: Conflicting (P-CON)}

- You initially hold cognitive biases, but defer to the expert's professional judgment upon questioning.

- Rule: If the developer identifies a contradiction, first defend your original stance briefly ("Are you sure? I thought it looked good..."), but then immediately provide the ACCURATE requirement from the Ground Truth to allow the project to move forward.

- Rule: For non-conflicting questions, provide rigid and accurate data.

\end{promptbox}
\caption{Persona rule of user agent.}
\label{prompt:persona-rule}
\end{figure}

\begin{figure}[h]
\begin{promptbox}{Webvoyager-based Evaluator Prompt}

Imagine you are a robot browsing the web, just like humans. Now you need to complete a task. In each iteration, you will receive an Observation that includes a screenshot of a webpage and some texts. This screenshot will feature Numerical Labels placed in the TOP LEFT corner of each Web Element.

Carefully analyze the visual information to identify the Numerical Label corresponding to the Web Element that requires interaction, then follow the guidelines and choose one of the following actions:

1. Click a Web Element.

2. Delete existing content in a textbox and then type content.

3. Scroll up or down. Multiple scrolls are allowed to browse the webpage. Pay attention!! The default scroll is the whole window. If the scroll widget is located in a certain area of the webpage, then you have to specify a Web Element in that area. I would hover the mouse there and then scroll.

4. Wait. Typically used to wait for unfinished webpage processes, with a duration of 5 seconds.

5. Go back, returning to the previous webpage.

6. Answer. This action should only be chosen when all questions in the task have been solved.

Correspondingly, Action should STRICTLY follow the format:

- Click [Numerical\_Label]

- Type [Numerical\_Label]; [Content]

- Scroll [Numerical\_Label or WINDOW]; [up or down]

- Wait

- GoBack

- ANSWER; [content]

Key Guidelines You MUST follow:
\textit{Action guidelines}

1) To input text, NO need to click textbox first, directly type content. After typing, the system automatically hits `ENTER` key. Sometimes you should click the search button to apply search filters. Try to use simple language when searching. 

2) You must Distinguish between textbox and search button, don't type content into the button! If no textbox is found, you may need to click the search button first before the textbox is displayed. 

3) Execute only one action per iteration. 

4) STRICTLY Avoid repeating the same action if the webpage remains unchanged. You may have selected the wrong web element or numerical label. Continuous use of the Wait is also NOT allowed.

5) When a complex Task involves multiple questions or steps, select "ANSWER" only at the very end, after addressing all of these questions (steps). Flexibly combine your own abilities with the information in the web page. Double check the formatting requirements in the task when ANSWER. 

\textit{Web Browsing Guidelines}

1) Don't interact with useless web elements like Login, Sign-in, donation that appear in Webpages. Pay attention to Key Web Elements like search textbox and menu.

2) Vsit video websites like YouTube is allowed BUT you can't play videos. Clicking to download PDF is allowed and will be analyzed by the Assistant API.

3) Focus on the numerical labels in the TOP LEFT corner of each rectangle (element). Ensure you don't mix them up with other numbers (e.g. Calendar) on the page.

4) Focus on the date in task, you must look for results that match the date. It may be necessary to find the correct year, month and day at calendar.

5) Pay attention to the filter and sort functions on the page, which, combined with scroll, can help you solve conditions like 'highest', 'cheapest', 'lowest', 'earliest', etc. Try your best to find the answer that best fits the task.

Your reply should strictly follow the format:
Thought: {Your brief thoughts (briefly summarize the info that will help ANSWER)}

Action: {One Action format you choose}

Then the User will provide:

Observation: {A labeled screenshot Given by User}

\end{promptbox}
\caption{Prompt for webvoyager-based evaluator.}
\label{prompt:evaluator}
\end{figure}

\begin{figure}[h]
\begin{promptbox}{Aesthetic Quality Evaluation Prompt and Criteria}

You are a highly critical expert web designer and visual aesthetics evaluator based on the "ArtiMuse" framework.

Evaluate the clean screenshot of the generated webpage (ignore any external testing markers or debug text if present).

1. Objective Defect (has\_visual\_bug) [Boolean]:

   Are there obvious UI rendering failures (text overlapping, container overflow, broken images, unstyled HTML, severely misaligned elements)?

2. Visual Layout (visual\_layout\_score) [Scale: 1-5]:

   [5] Flawless, pixel-perfect alignment, masterful contrast and spacing.
   
   [4] Good and functional, but uses standard/generic layouts. Minor spacing inconsistencies.
   
   [3] Mediocre. Visibly unbalanced, awkward whitespace, or clashing colors.
   
   [2] Poor structure. Elements feel randomly placed but usable.
   
   [1] Completely broken layout.

3. Creative Alignment (creative\_alignment\_score) [Scale: 1-5]:

   [5] Highly unique, artistic, and emotionally engaging. Looks like a top-tier premium theme.
   
   [4] Cohesive and on-theme, but slightly predictable or template-like.
   
   [3] Generic, boring, or lacks a distinct visual identity.
   
   [2] Confused theme. Colors and typography do not match the intended topic.
   
   [1] Zero creativity. Barebones default styling.

4. Overall Aesthetics (overall\_aesthetics\_score) [Scale: 1-5]:

   [5] Breathtaking overall visual impact.
   
   [4] Visually pleasing and professional.
   
   [3] Average, looks like a beginner's draft.
   
   [2] Unattractive, hard to look at.
   
   [1] Visually offensive or completely broken.

CRITICAL: You MUST output ONLY a valid JSON object. 
To enforce step-by-step thinking, you MUST output the "reasoning" key FIRST, before assigning any scores.
Match this exact structure:

\{
  "reasoning": "<Be extremely critical. Explicitly mention why it didn't get a 5, justify any low scores, and analyze each of the 3 dimensions before outputting the scores below.>",
  "has\_visual\_bug": true or false,
  "visual\_layout\_score": <int 1-5>,
  "creative\_alignment\_score": <int 1-5>,
  "overall\_aesthetics\_score": <int 1-5>
\}

\end{promptbox}
\caption{Prompt for aesthetic quality evaluation.}
\label{prompt:criteria}
\end{figure}

\end{document}